%% file: main.tex
\documentclass{article}

\usepackage{microtype}
\usepackage{graphicx}
\usepackage{subfigure}
\usepackage{booktabs} %
\usepackage{algorithm}
\usepackage{algorithmic}
\usepackage{multirow}
\usepackage{tabularx}
\usepackage{enumerate}
\usepackage{lipsum}

\usepackage{hyperref}

\usepackage[accepted]{icml2023}

\usepackage{amsmath}
\usepackage{amssymb}
\usepackage{mathtools}
\usepackage{amsthm}

\DeclareMathOperator{\subc}{\mathcal{C}}
\DeclareMathOperator{\ego}{ego}
\DeclareMathOperator{\diam}{diam}
\DeclareMathOperator{\pert}{\mathcal{P}}
\DeclareMathOperator{\argmax}{argmax}

\usepackage{acronym}
\newacro{GNN}{Graph Neural Netwok}
\acrodefplural{GNN}[GNNs]{Graph Neural Netwoks}
\newacro{MLP}{Multi-Layer Percepton}
\acrodefplural{MLP}[MLPs]{Multi-Layer Perceptons}
\newacro{MPNN}{Message Passing Neural Network}
\acrodefplural{MPNN}[MPNNs]{Message Passing Neural Networks}
\newacro{WL}{Weisfeiler-Lehman}
\newacro{AUC}{area under the curve}
\newacro{OOD}{out-of-distribution}

\usepackage[capitalize,noabbrev]{cleveref}

\theoremstyle{plain}
\newtheorem{theorem}{Theorem}[section]
\newtheorem{proposition}[theorem]{Proposition}

\theoremstyle{definition}
\newtheorem{definition}[theorem]{Definition}

\theoremstyle{remark}

\usepackage[textsize=tiny]{todonotes}

\icmltitlerunning{Expressivity of Graph Neural Networks Through the Lens of Adversarial Robustness}

\begin{document}

\twocolumn[
\icmltitle{Expressivity of Graph Neural Networks \\ Through the Lens of Adversarial Robustness}

\begin{icmlauthorlist}
\icmlauthor{Francesco Campi}{TUM}
\icmlauthor{Lukas Gosch}{TUM}
\icmlauthor{Tom Wollschläger}{TUM}
\icmlauthor{Yan Scholten}{TUM}
\icmlauthor{Stephan Günnemann}{TUM}
\end{icmlauthorlist}

\icmlaffiliation{TUM}{TUM School of Computation, Information and Technology \& Munich Data Science Institute, Technical University of Munich, Germany}

\icmlcorrespondingauthor{Francesco Campi}{francesco.campi98@gmail.com}

\icmlkeywords{Machine Learning, ICML}

\vskip 0.3in
]

\printAffiliationsAndNotice{%
} %

\begin{abstract}
We perform the first adversarial robustness study into Graph Neural Networks (GNNs) that are provably more powerful than traditional Message Passing Neural Networks (MPNNs). In particular, we use adversarial robustness as a tool to uncover a significant gap between their theoretically possible and empirically achieved expressive power. To do so, we focus on the ability of GNNs to count specific subgraph patterns, which is an established measure of expressivity, and extend the concept of adversarial robustness to this task. Based on this, we develop efficient adversarial attacks for subgraph counting and show that more powerful GNNs fail to generalize even to small perturbations to the graph's structure. Expanding on this, we show that such architectures also fail to count substructures on out-of-distribution graphs.

\end{abstract}

\input{include/1-introduction}

\input{include/2-background}
\input{include/3-related_work}

\input{include/4-adv_robustness}
\input{include/5-adv_attack}
\input{include/6-experiments}

\input{include/7-conclusion}

\bibliography{example_paper}
\bibliographystyle{icml2023}

\newpage
\appendix
\onecolumn
\input{include/Appendix}

\end{document}

%% file: include/1-introduction.tex
\section{Introduction}
In recent years, significant efforts have been made to develop \acp{GNN}, for several graph-related tasks, such as molecule property predictions \citep{Dimenet}, social network analysis \citep{Socialrec}, or combinatorial problems \citep{Gasse2019ExactCO}%
, to name a few. 
The most commonly used architectures are based on message passing, which iteratively updates the embedding of each node based on the embeddings of its neighbors \citep{MPNN}. Despite their broad success and %
wide adoption, different works have pointed out that so called \acp{MPNN} are at most as powerful as the 1-\ac{WL} algorithm \citep{morris_weisfeiler_2021, how_Powerful} and thus, have important limitations in their expressive power \citep{chen_can_2020}. 
\begin{figure}
    \centering
    \includegraphics[width=\columnwidth]{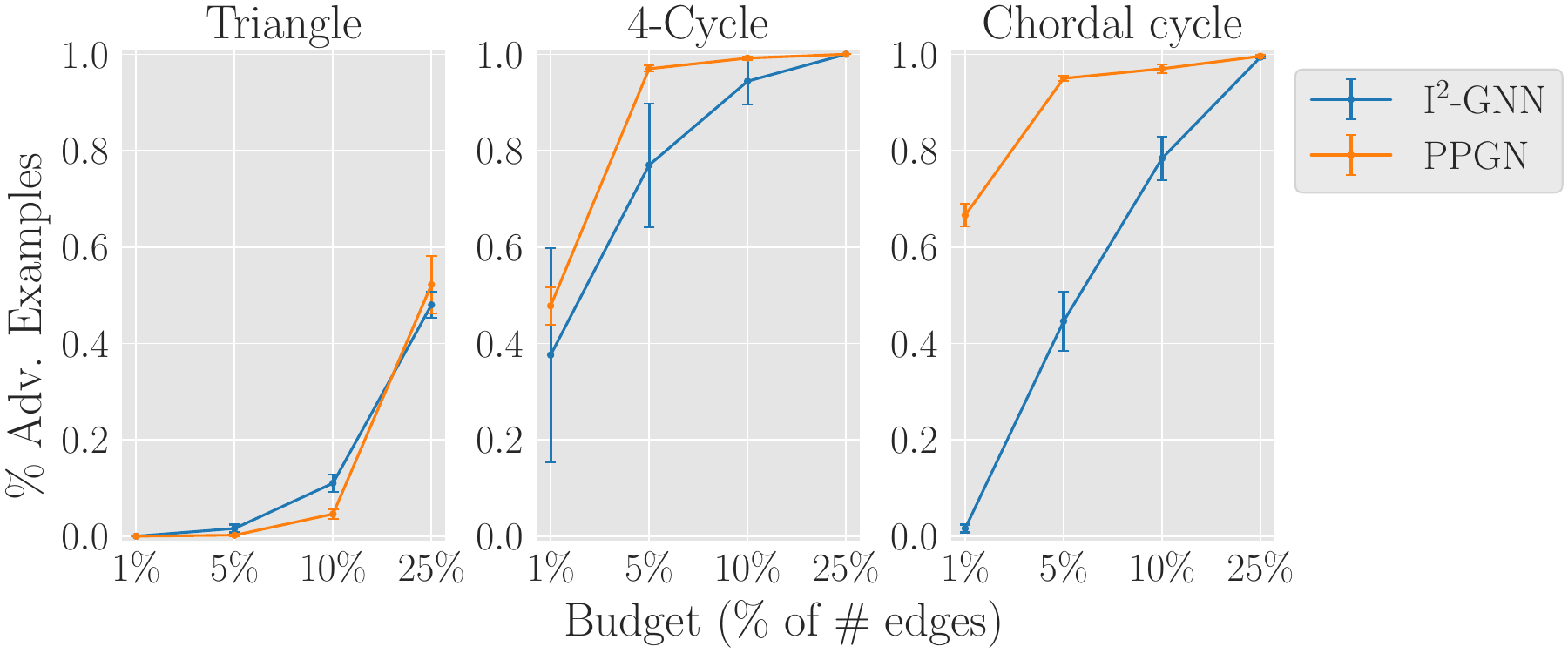}
    \caption{\acp{GNN} more powerful than 1-\ac{WL} are not adversarially robust for subgraph-counting tasks.}
    \label{fig:rob intro}
\end{figure}
This encouraged the development of provably more powerful architectures.
\par
However, there is no guarantee that the training process also yields models that are as powerful as theoretically guaranteed.
Thus, this work investigates if and to what extent the empirically achieved expressivity of such GNNs lacks behind their theoretic possibilities by taking a novel look from the perspective of adversarial robustness.
In particular, we focus on the task of counting different subgraphs, which is provably impossible for \acp{MPNN} \citep{chen_can_2020} (except for very limited cases), but important for many downstream tasks \citep{huang_boosting_2022, liu2019ngram, motifnet_2018}. %
\par
Using our new adversarial framework for subgraph counting, we find that the counting ability of theoretically more powerful GNNs fail to generalize even to small perturbations to the graph's structure (see \cref{fig:rob intro}). A result even more interesting given that subgraph counting is polynomially solvable for fixed subgraph sizes \citep{shervashidze_ecient_nodate}.%
\footnote{In general, this problem is NP-complete \citep{ribeiro_survey_2022}}
We expand on these results and show that these architectures also fail to count substructures on \ac{OOD} graphs. Furthermore, retraining the last MLP layers responsible for the prediction based on the graph embedding does not entirely resolve this issue.
\par
\textbf{Contributions.} (i) We perform the first study into the adversarial robustness of \acp{GNN} provably more powerful than the 1-WL and use it as an effective tool to uncover a significant gap between the theoretically possible and empirically achieved expressivity for substructure-counting tasks (see \Cref{sec:experiments}).
(ii) We extend the concept of an adversarial example from classification to (integer) regression tasks and develop multiple perturbations spaces interesting for the task of subgraph counting (see \Cref{sec: adv rob}). 
(iii) We develop efficient and effective adversarial attacks for subgraph counting, operating in these perturbations spaces and creating \emph{sound} perturbations, i.e., where we know the ground truth (see \Cref{sec:adv att}).
(iv) In \Cref{sec:exp_ood} we show that subgraph-counting GNNs also fail to generalize to out-of-distribution graphs, providing additional evidence that these GNNs fail to reach their theoretically possible expressivity. %
Our code implementation can be found at \href{https://github.com/francesco-campi/Rob-Subgraphs}{https://github.com/francesco-campi/Rob-Subgraphs}.

%% file: include/2-background.tex
\section{Background}
We consider undirected, unattributes graphs~\(G\)=\((V, E)\) with nodes $V$=$\{1,\ldots, n\}$ and edges \(E\)\(\subseteq\)\(\{ \{i,j\} \mid i,j \in V, i\)\(\neq\)\(j \}\), represented by adjacency matrix $\mathbf{A} \in \{0,1\}^{n \times n}$. 
A graph $G_S = (V_S, E_S)$ is a \emph{subgraph} of $G$ if $V_S \subseteq V$ and $E_S \subseteq E$. 
We say $G_S$ is an \emph{induced} subgraph if $E_S$ contains all edges in $E$ that connect pairs of nodes in $V_S$.
An egonet $\ego_l(i)$ is the induced subgraph containing all nodes with a distance of at most $l$ from root node $i$. 
Furthermore, two graphs $G, G'$ are isomorphic ($\simeq$) if there exists a bijection $f: V \to V'$ such that $\{i,j\} \in E$ if and only if $\{f(i), f(j)\} \in E'$.
Lastly, the diameter $\diam(G)$ denotes the length of the largest shortest path in graph $G$.

\subsection{Subgraph-Counting}
\begin{figure}
    \centering
    \begin{tabular}{cccc}
        \centering
         {\includegraphics[width=0.12\columnwidth]{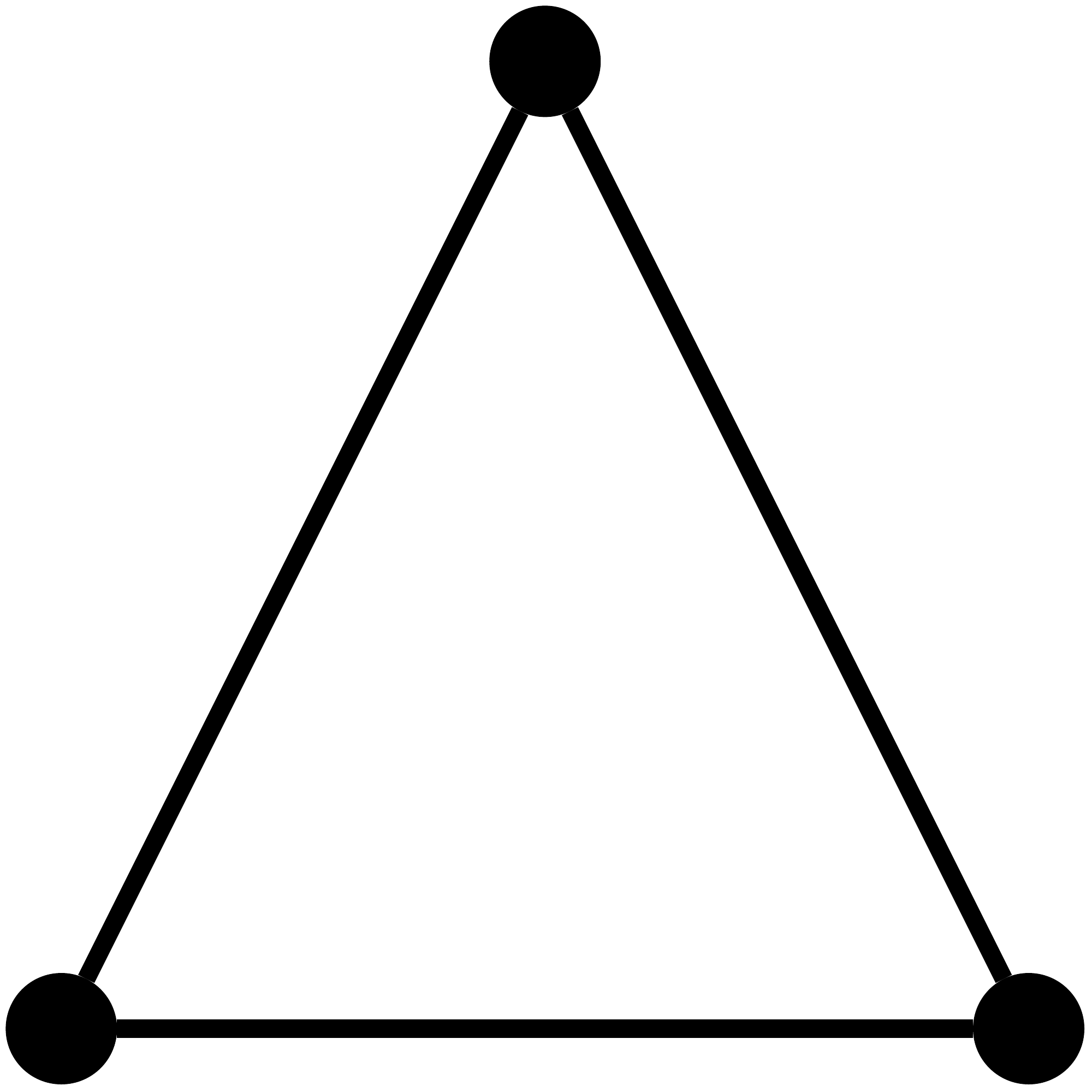}}&{\includegraphics[width=0.12\columnwidth]{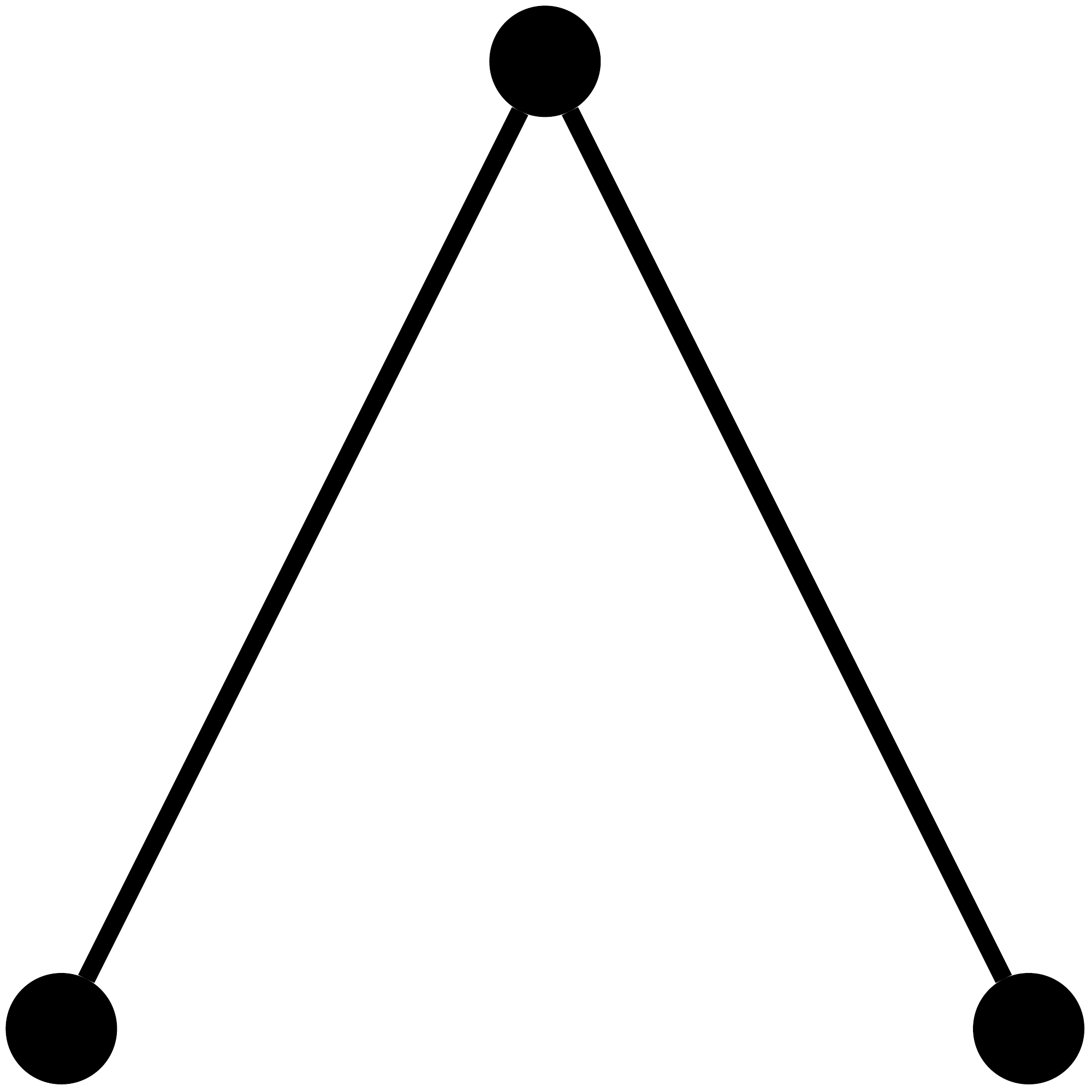}}&\includegraphics[width=0.12\columnwidth]{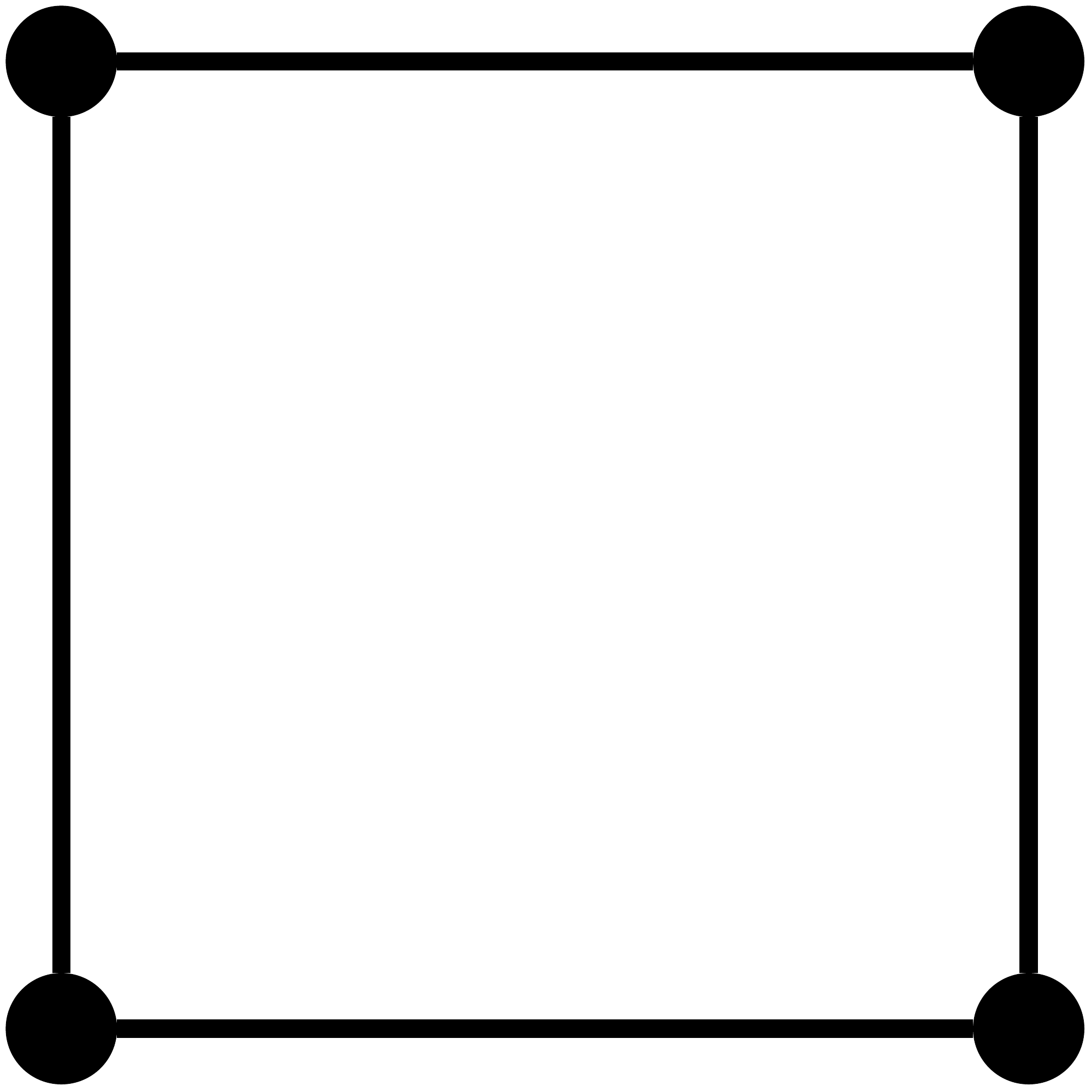}&{\includegraphics[width=0.12\columnwidth]{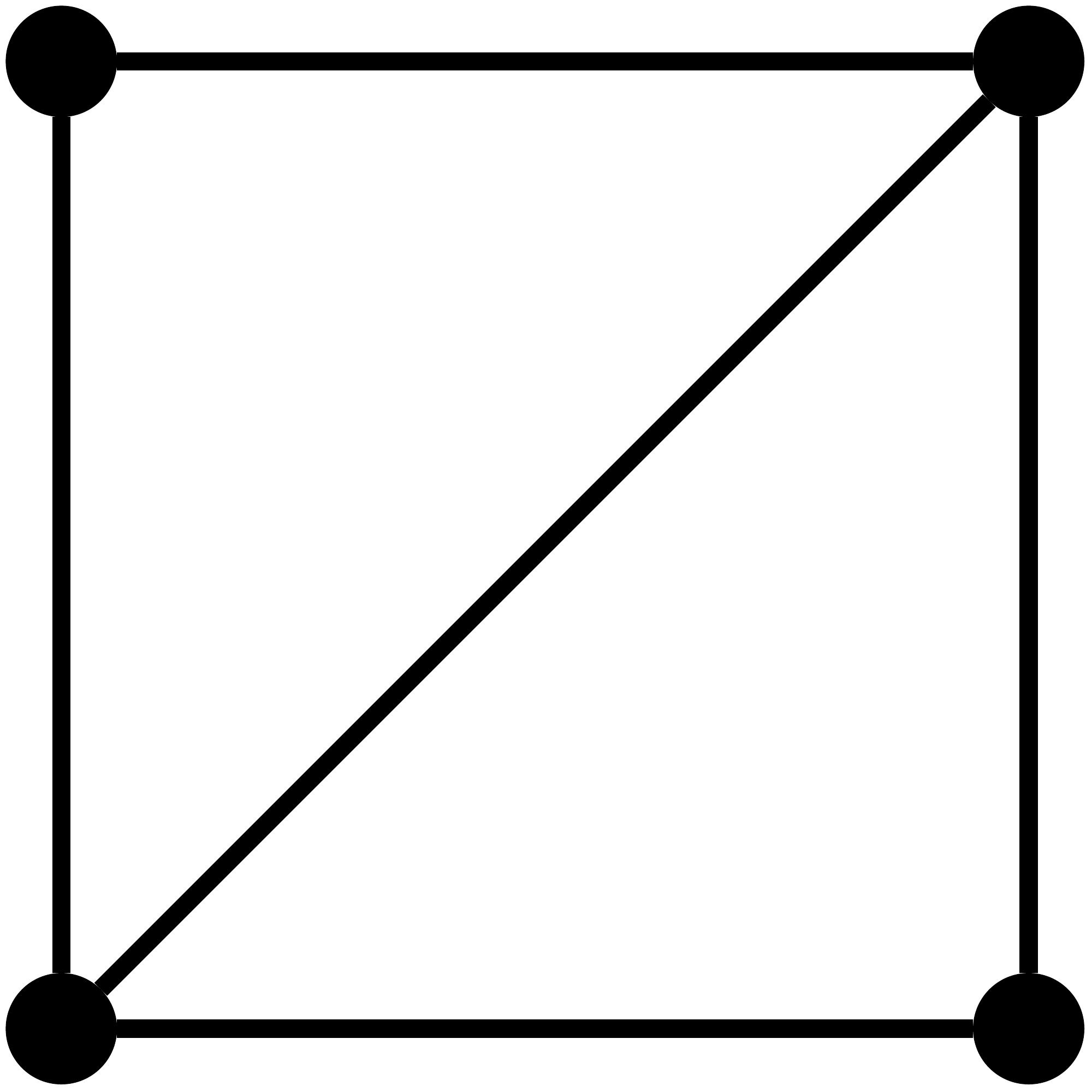}}\\
         \small Triangle & \small 2-Path&\small 4-Cycle&\small Chordal Cycle
    \end{tabular}
    \caption{Examples of graph patterns used for subgraph-counting.}
    \label{fig:pattern}
\end{figure}

Consider a fixed %
graph $H$ which we call a pattern (\cref{fig:pattern}).
A classic graph-related problem is the \emph{(induced-) subgraph-counting} of the pattern $H$ \citep{ribeiro_survey_2022}, which consists of enumerating the (induced) subgraphs of $G$ isomorphic to $H$. The subgraph-count of $H$ is denoted by $\subc(G, H)$, and by $\subc_I(G, H)$ in the induced case. To simplify the notation we will also refer to it as $\subc(G)$ if $H$ is given in the context. 
Several algorithms have been developed to solve the task of subgraph-counting. In this work we specifically consider the (exact) algorithm of \citet{shervashidze_ecient_nodate} (presented in \cref{sec:counting alg}) due to its low computational cost. %

\subsection{Expressivity of Graph Neural Networks}
The expressivity of machine learning models is about which functions they can and cannot approximate. %
There are different ways of studying the expressive power of \acp{GNN}. In this work we specifically consider their ability to count subgraphs \citep{chen_can_2020} because it is strictly related to different real-world tasks such as computational chemistry \citep{Jin2020ComposingMW} and social network analysis \citep{social_net_sub}. We define the ability to count subgraphs as~follows:

\begin{definition}
A family of functions $\mathcal{F}$ can perform \emph{subgraph-counting} of a target pattern $H$ on a graph class $\mathcal{G}$ if for any two graphs $ G_1, G_2 \in \mathcal{G}$ with $\subc(G_1, H) \neq \subc(G_2, H)$ there exists a function $f \in \mathcal{F}$ such that $f(G_1) \neq f(G_2)$.
\label{def:subgraph_counting}
\end{definition}
\begin{figure}[t]
    \centering
    \includegraphics[width=0.17\textwidth]{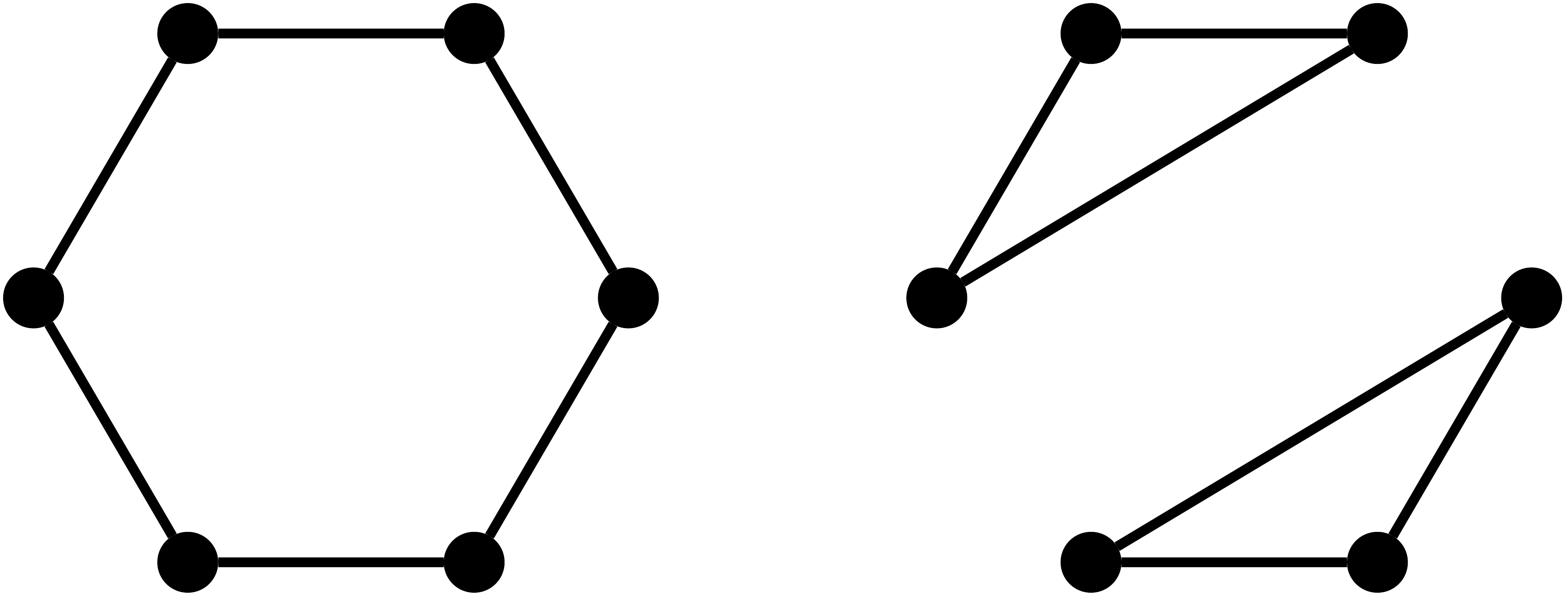}
    \caption{Pair of undistinguishable graphs for \acp{MPNN} with different triangle counts.}
    \label{fig:undistinguishable}
\end{figure}
Surprisingly, \acp{MPNN} have considerable limitations in subgraph-counting. In fact, \citet{chen_can_2020} show that \acp{MPNN} are not able to count induced patterns with three or more nodes, leaving out only the ability to count edges. For example, \cref{fig:undistinguishable} shows two graphs that, despite having different triangle counts, will always return identical outputs when fed in the same \ac{MPNN}.

A different perspective to measure the expressive power is graph isomorphism. In this regard, \citet{how_Powerful, morris_weisfeiler_2021} demonstrated that an \ac{MPNN} is at most as powerful as 1-\ac{WL} isomorphism test at distinguishing pairs of non-isomorphic graphs. Moreover, since the \ac{WL} algorithms are designed to extract representation vectors from graphs, they could be used also to perform subgraph-counting.
In particular, \citet{chen_can_2020} showed that $k$-\ac{WL}, and equivalently powerful architectures,  can perform substructure-counting for patterns with at most $k$ nodes, creating a connection between the two approaches.

\subsection{More Expressive Graph Neural Networks}

In this work, we analyze two state-of-the-art architectures for the task of subgraph counting: PPGN \citep{maron_provably_2020}, and I$^2$-GNN \citep{huang_boosting_2022}. PPGN represents the graph structure in a tensor and exploits tensor multiplications to enhance the expressivity. It reaches the same expressive power of 3-\ac{WL}, which makes it capable of counting patterns of size three.
I$^2$-GNN, following the approach of subgraph GNNs \citep{frasca2022understanding}, decomposes the whole graph into different subgraphs and processes them independently with an \ac{MPNN}. It has been explicitly developed to be expressive enough for counting different substructures and most important for this work, can count arbitrary patterns of size four. Both, PPGN and I$^2$-GNN are effective architectures for downstream tasks such as molecular property~predictions. 

%% file: include/3-related_work.tex
\section{Related Work}
\citet{chen_can_2020} were the first to study the expressivity of GNNs w.r.t.\ their ability to count substructures. They, and later \citet{tahmasebi_counting_2021} proposed architectures for counting substructures. However, these suffer from high computational complexity. \citet{yu2023learning} proposed an architecture purely focusing on subgraph counting. However, subgraph counting alone can be solved by efficient randomized algorithms \citep{bressan}. Thus, in this work, we focus on efficient architectures, which leverage their subgraph counting ability to improve generalization for other downstream tasks. In particular, we focus on PPGN \citep{maron_invariant_2019} and I$^2$-GNN \citep{huang_boosting_2022}. Both achieve state-of-the-art results for substructure counting while having formal expressivity guarantees. %

Different works have studied the adversarial robustness of \acp{GNN} for graph-level classification \citep{dai_adversarial_2018} and node-level classification \citep{zugner_adversarial_2018}. Regarding the latter, \citet{gosch2023revisiting} exactly define (semantic-preserving) adversarial examples. Moreover, \citet{geisler_generalization_2022} use adversarial attacks with sound perturbation models, i.e., where the ground truth change is known, to investigate the generalization of neural combinatorial solvers. Conversely, adversarial robustness for regression tasks has currently received very little attention \citep{deng_analysis_2020}.

%% file: include/4-adv_robustness.tex
\section{Robustness in Subgraph-Counting}
\label{sec: adv rob}
The field of adversarial robustness is about the problem that machine learning models are vulnerable to small changes to their inputs \cite{goodfellow2015explaining}. In particular, for the subgraph-counting problem we want to analyze whether the error of the models increases when tested on perturbed input graphs $\Tilde{G}$ of a graph from a set of perturbed graphs $\pert(G)$. To evaluate the performance of a model $f$ on perturbed graphs $\Tilde{G}\in\pert(G)$ we use the following adversarial loss:
\[
    \ell_{adv}(\Tilde{G}) := |f(\Tilde{G}) - \subc(\Tilde{G}, H)|.
\]

\subsection{Subgraph-Counting Adversarial Examples}
To empirically evaluate the expressivity of machine learning models for subgraph-counting via adversarial robustness, we have to introduce a notion of adversarial example.
In classification tasks adversarial examples are simply perturbations that change the predicted class. 
In general regression tasks %
one can define a threshold on \(\ell_{adv}\) for which we call a perturbed graph an adversarial example  \cite{deng_analysis_2020}. However, this definition is application-dependent and, in our work, we define a specific threshold exploiting the fact that subgraph-counting is an \emph{integer} regression task. %
\begin{definition}
\label{def:adv example}
    Given a model $f$ and clean graph $G$, we say that $\Tilde{G} \in \pert(G)$ is an \emph{adversarial example} for $f$ if:
    \begin{enumerate}[(i)]
        \item  $\lfloor f(G) + 0.5\rfloor = \subc(G)$
        \item  $\lfloor f(\Tilde{G}) + 0.5\rfloor \neq \subc(\Tilde{G})$
        \item $\dfrac{\ell_{adv}(\Tilde{G}) - \ell_{adv}(G)}{\ell_{adv}(G)} > \delta$. 
        \label{def:adv example iii}
    \end{enumerate}
\end{definition}
The conditions $(i)$ and $(ii)$ guarantee that the model prediction, when approximated to the nearest integer, is correct for $G$ and wrong for $\Tilde{G}$. Here, having a correct initial prediction is essential to clearly distinguish the performances on the original graph from the perturbed graph.
In addition, the condition $(iii)$ ensures that a margin exists between the errors on the original data instance and the perturbed one, and the size of the margin depends on the value of $\delta$. This requisite prevents easily generating adversarial examples from graphs that are almost wrongly predicted, i.e. $\ell_{adv}(G) \approx 0.5$.

\subsection{Perturbation Spaces}
We define different perturbation spaces for a graph $G$ as constrained sets of structurally perturbed graphs constructed from $G$%
. In particular, we consider different combinations of edge deletions and additions, for example $E' = E \cup \{i,j\}$ with $\{i,j\} \notin E$ represents an edge addition. %
We always consider sound perturbation models, i.e, where we know the ground truth change. These are efficiently implemented as described in \cref{sec:adv att}. %
It is meaningful to limit the number of perturbations in order to control how shifted the distribution of the perturbed subgraph-counts is compared to the distribution of the original ones.
Then, we define the \emph{constrained} perturbation space with maximal budget $\Delta$ as:
\begin{equation}
    \pert_\Delta (G) \coloneqq \{\Tilde{G} \ | \ \frac{1}{2}\|\mathbf{A} - \mathbf{A'}\|_0 \le \Delta\},
    \label{eq:pert}
\end{equation}
where $\|\cdot\|_0$ represents the number of non-zero elements, i.e.\ the number of perturbed edges.
\par
\textbf{Semantic-Preserving Perturbations.} Additionally, we conduct a robustness analysis more closely in line with adversarial examples for classification tasks, by incorporating a further constraint to guarantee the preservation of a specific level of semantic meaning. %
In particular, we define the \emph{count-preserving} perturbation space as:
\begin{equation}
    \pert_\Delta^c (G) \coloneqq \{\Tilde{G} \ | \ \Tilde{G} \in \pert_\Delta(G) \ \wedge \ \subc(\Tilde{G}) = \subc(G)\}.
    \label{eq:count_pert}
\end{equation}
\par
\begin{figure}
    \centering
    \includegraphics[width=0.27\textwidth]{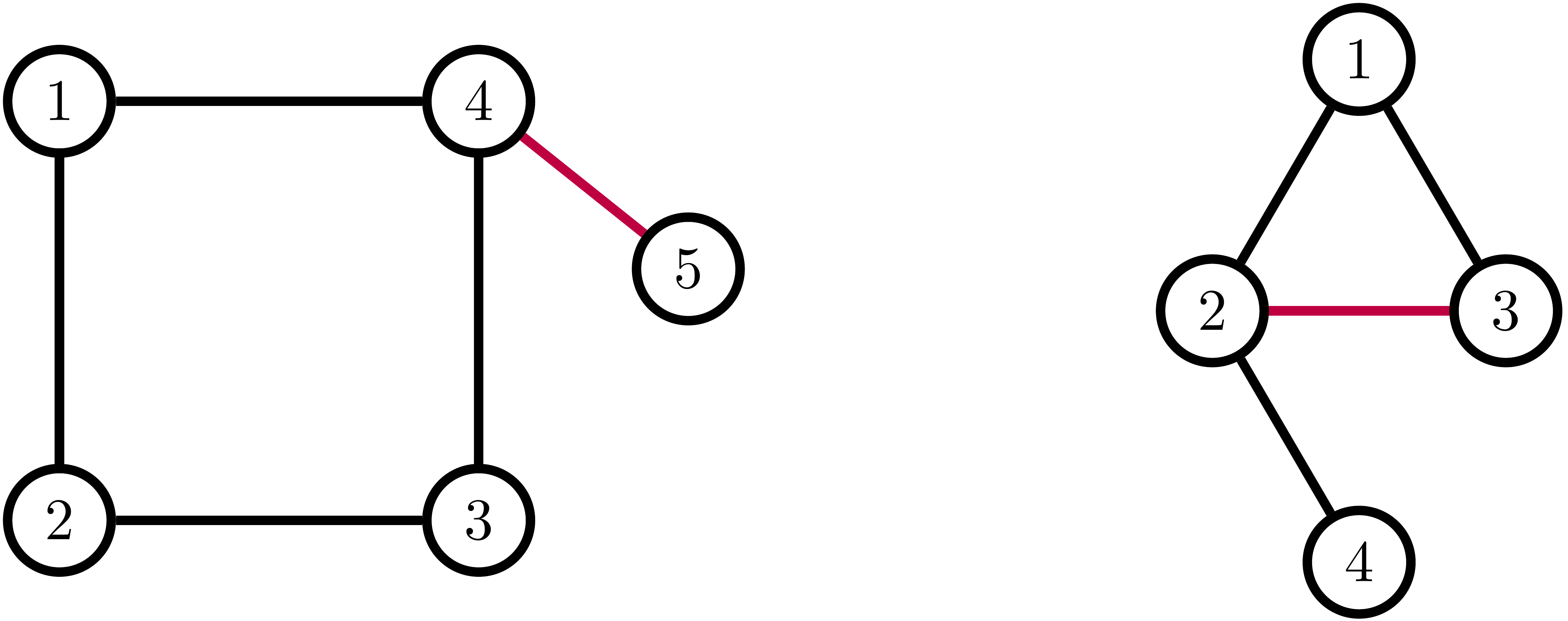}
    \caption{This Figure shows examples demonstrating that not all the count-preserving perturbations are also subgraph-preserving ones. On the left a subgraph- and count-preserving perturbation for 4-cycles where the red edge has been deleted. On the right a perturbation that leaves unchanged the count of 2-paths, but it deletes the induced substructure $\{2,3,4\}$ to generate $\{1,2,3\}$.}
    \label{fig:perturbation}
\end{figure}

Additionally, when considering induced subgraphs, keeping the count constant does not guarantee that the subgraphs isomorphic to the pattern remain the same. In fact, perturbations can simultaneously delete a subgraph isomorphic to the pattern and generate a new one (see \cref{fig:perturbation}). We will denote the \emph{subgraph-preserving} perturbation space by 
\begin{equation}
\begin{split}
   \pert^s_\Delta&(G) \coloneqq \{\Tilde{G} \ | \ \Tilde{G} \in \pert_\Delta(G) \ \wedge \\ & G_S \subseteq G, G_S \simeq H \iff G_S \subseteq \Tilde{G}, G_S \simeq H\}. 
\end{split}
\end{equation}

%% file: include/5-adv_attack.tex
\section{Subgraph-Counting Adversarial Attacks}
\label{sec:adv att}
For a subgraph-counting model $f$, the goal of an adversarial attack is to find the pertubed graph $G^*\in\pert(G)$ that causes the maximal error increase. This problem can be formulated as an optimization problem:
\begin{equation}
    G^* = \underset{\Tilde{G} \in \pert(G)}{\argmax} \; \ell_{adv}(\Tilde{G}).
    \label{eq:adv attack}
\end{equation}
Attacking subgraph-counting \acp{GNN} for studying their empirical expressivity is particularly challenging. In fact, (i) the subgraph-count can vary significantly even for slight structural changes, and (ii) finding $G^*$ of \cref{eq:adv attack} requires solving a discrete optimization problem.

\subsection{Sound Perturbations for Subgraph-Counting}
\label{sec:groud update}
 To tackle the sensitivity of the counts to structural changes, we exploit the exact algorithm to update the ground-truth count after every perturbation. In this way, we generate sound perturbations since the exact ground-truth value is know. In order to prevent this step to become computationally prohibitive,  we develop an efficient count updating scheme that uses only a small portion of the graph.
\begin{proposition}
    Consider a graph $G$ and a pattern $H$ with $\diam(H) = d$. Then, for every edges $\{i,j\}$ we have that $\ego_d(i)$ and $\ego_d(j)$ contain all the subgraphs $G_S \subset G$ such that $G_S \simeq H$ and $i,j \in V_S$.
    \label{thm:ego_pert}
\end{proposition}%
Proof in \cref{sec:thm proof}.

\begin{algorithm}
   \caption{Beam search (greedy search for $k=1$)}
   \label{alg:greedy}
\begin{algorithmic}
   \STATE {\bfseries Input:} $G, \Delta, k$
   \STATE $\mathcal{G}^{(0)} = \{G\}$
   \FOR{$i=0$ {\bfseries to} $\Delta -1$}
   \STATE $\pert^{(i)} = \{\}$
   \FOR{$\Tilde{G}$ {\bfseries in} $\mathcal{G}^{(i)}$}
   \STATE $\pert^{(i)} = \pert^{(i)}\cup \pert_1(\Tilde{G})$
   \hfill\COMMENT{or $\pert_1^c(G)$, $\pert^s_1(G)$}
   \ENDFOR
   \STATE $\mathcal{G}^{(i+1)} = \text{greatest } k \text{ in } \{\ell_{adv}(\Tilde{G}) \ | \ \Tilde{G} \in \pert^{(i)}\}$
   \ENDFOR
   \STATE {\bfseries Return: } $G^* = \argmax_{\Tilde{G} \in \mathcal{G}^{(\Delta)}} \{\ell_{adv}(\Tilde{G})\}$
\end{algorithmic}
\end{algorithm}

When an edge $\{i,j\}$ is perturbed, only the subgraphs containing both the end nodes can be affected and potentially change their isomorphism relation with $H$. Therefore, according to \cref{thm:ego_pert}, it is sufficient to verify potential count changes only in $\ego_d(i)$ (or equivalently $\ego_d(j)$). Specifically, the theorem assumes that $\{i,j\}$ is contained in the graph, hence we extract the egonet from the graph including $\{i,j\}$ (original for edge deletion and perturbed for addition). Next, from the nodes of $\ego_d(i)$ we generate the induced subgraphs $G_S$ and $\Tilde{G}_S$ from the original and perturbed graphs respectively. Since the possible alterations of the subgraph-count are enclosed in $G_S$ and $\Tilde{G}_S$, we have the following count update rule.
\begin{proposition}
    Let $\Tilde{G}$ be a perturbation of a single edge of a graph $G$, then there holds:
    \label{eq:count up}
    \[\subc(\Tilde{G}) = \subc(G) + \subc(\Tilde{G}_S)- \subc(G_S).\]
\end{proposition}
Following \cref{eq:count up} we need to run the subgraph-counting algorithm only on the smaller subgraphs $G_S$ and $\Tilde{G}_S$, rather than on the whole graph $\Tilde{G}$.
Additionally, \cref{thm:ego_pert}
guarantees that potential changes in the subgraphs isomorphic to the patterns are also constrained in the egonet, thus it can be used also identify perturbations belonging to the subgraph-preserving perturbation space $\pert_\Delta^s$ .

\subsection{Construction of Adversarial Examples}
To create adversarial examples we need to solve the discrete optimization problem  in \cref{eq:adv attack}. To do so we develop algorithms that generate more powerful perturbation one change at a time, in this way, we keep track of the exact count with the update rule (\cref{eq:count up}).

\textbf{Greedy Search.} We develop an efficient and effective greedy search algorithm (\cref{alg:greedy}).
At each step we select the most effective perturbation of the current perturbed graph $\Tilde{G}$ in $\pert_1(\Tilde{G})$ (or in $\pert_1^c(\Tilde{G}), \pert_1^s(\Tilde{G})$) until the budget limit is reached. 
The new subgraph-counts of perturbations in $\pert_1(\Tilde{G})$ are computed  with \cref{eq:count up}, whereas the preserving perturbation spaces are generated with \cref{alg:pret sapce}.

\textbf{Beam search.} A more advanced algorithm that does not increase the computational complexity is beam search. Concretely, it follows simultaneously $k$ different paths to explore more extensively the perturbation space (see \cref{alg:greedy}).
\par
To improve the computational efficiency the perturbations in $\pert_1$ can be randomly selected according to the degrees of the end nodes of the perturbed edge. Concretely, the probability to pick the perturbation where the edge $\{i,j\}$ has been added (or deleted) is proportional to $d(i)^2 +d(j)^2$, since intuitively these are the most relevant edges.

\begin{algorithm}[t]
   \caption{$\pert_1^c(G)$ generation (analogous for $\pert_1^s(G)$)}
   \label{alg:pret sapce}
\begin{algorithmic}
   \STATE {\bfseries Input:} $G, \subc(G)$
   \STATE $\pert_1^c(G)= \{\}$
   \FOR{$\Tilde{G}$ {\bfseries in} $\pert_1(G)$}
   \STATE $\subc(\Tilde{G}) = \subc(G) + \subc(G_S')- \subc(G_S)$
   \IF{$\subc(\Tilde{G}) = \subc(G)$} 
   \STATE $\pert_1^c(G)= \pert_1^c(G) \cup\{\Tilde{G}\}$
   \ENDIF
   \ENDFOR
   \STATE {\bfseries Return: } $\pert_1^c(G)$
\end{algorithmic}
\end{algorithm}

%% file: include/6-experiments.tex
\section{Experiments} \label{sec:experiments}
\label{sec:exp}

In \cref{sec:exp adv rob}, we analyze the empirical expressivity of \acp{GNN} using our subgraph-counting adversarial attacks and using generalization as a (proxy) measure. Extending on this, in \Cref{sec:exp_ood} we investigate if the same GNNs can count subgraph patterns for out-of-distribution graphs. Here we present the results of the induced subgraph-counting of triangles, 4-cycles and chordal cycles, for other patterns refer to \cref{sec:add res}.

\subsection{Adversarial Robustness}
\label{sec:exp adv rob}
Here, we analyze the empirical expressivity of  \acp{GNN} using our subgraph-counting adversarial attacks. %
\par
\textbf{Dataset and models.} We generate a synthetic dataset of $5{,}000$ Stochastic-Block-Model graphs with 30 nodes divided into 3 different communities. The probabilities of generating edges connecting nodes within the same community are $[0.2, 0.3, 0.4]$, while the probability of generating edges between nodes of different communities is 0.1. We randomly split the dataset into training, validation, and test sets with percentages $30\%, 20\%, 50\%$. We then train PPGN \cite{maron_provably_2020} and I$^2$-GNN \cite{huang_boosting_2022}. %

\par
\textbf{Experimental Settings.} We train each model 5 times using different initialization seeds to prevent bad weight initialization influencing the final results. Then, for each of the trained models $f_i$ with seed $i$, we use our adversarial attacks (see \cref{sec:adv att}) to generate adversarial examples from 100 correctly predicted test graphs and average the percentage of successful attacks over all seeds. %
Furthermore, we investigate if the adversarial graphs for a model $f_i$ transfer 
to the other models $f_j$ trained with a different initialization seed $j \neq i$. We inspect all three different perturbation spaces with budgets $\Delta$ of $ 1\%, 5\%, 10\%$ and $ 25\%$ with respect to the average number of edges of the graphs in the dataset and use $\delta = 1$ as margin. 
In detail, we use beam search with beam width $k = 10 $ to explore $\pert^c_\Delta$ and $\pert^s_\Delta$, while we rely on greedy search for~$\pert_\Delta$.

\par
\begin{figure}[!ht]
    \centering
    \begin{subfigure}
    \centering
    \includegraphics[width=\columnwidth]{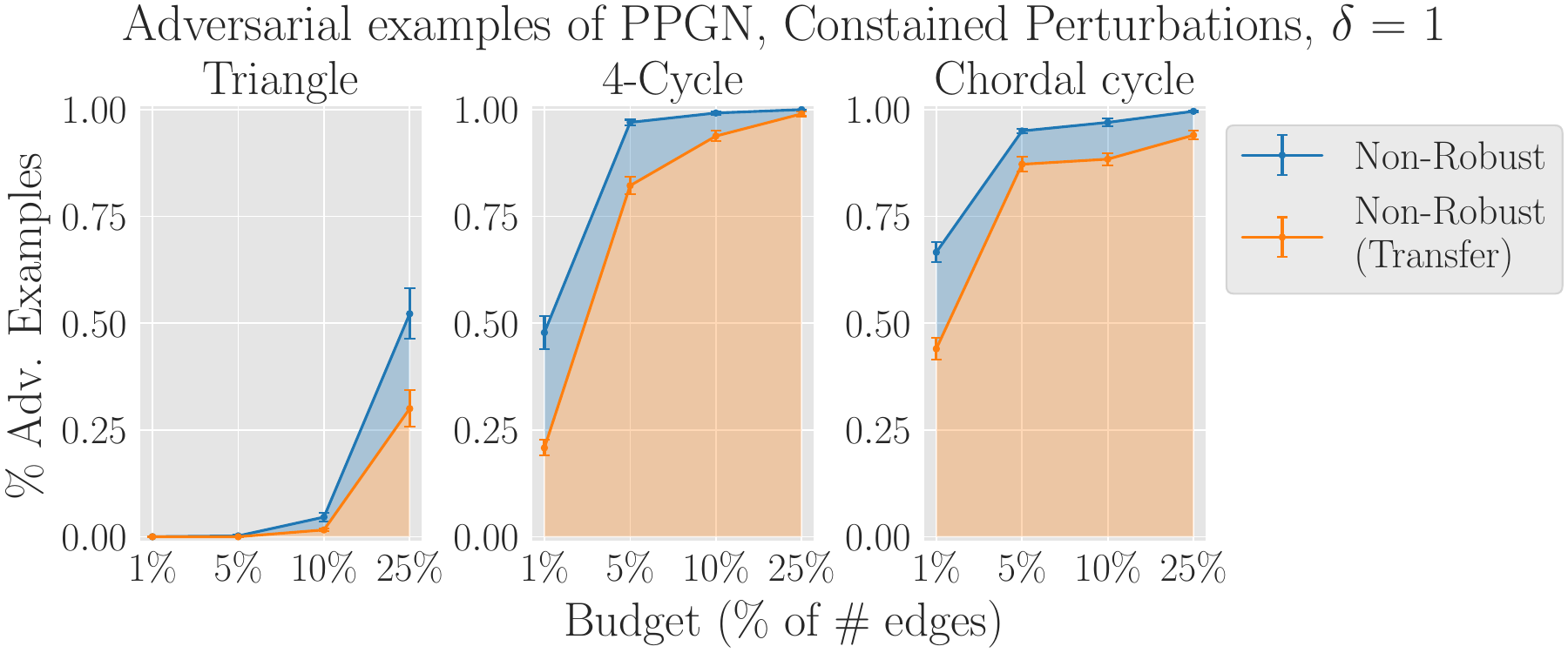}
    \end{subfigure}
    \begin{subfigure}
    \centering
    \includegraphics[width=\columnwidth]{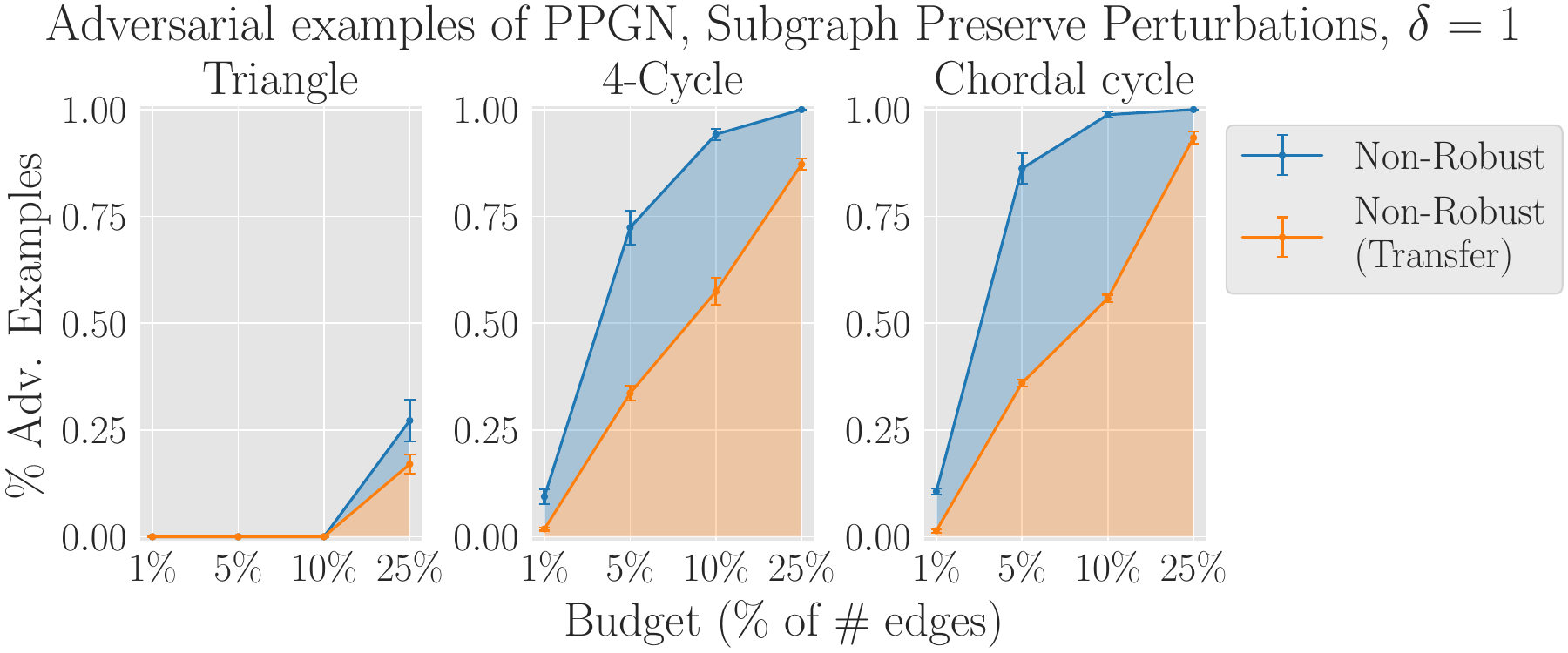}
    \end{subfigure}
    \vspace{0.3cm}
    
    \begin{subfigure}
    \centering
    \includegraphics[width=\columnwidth]{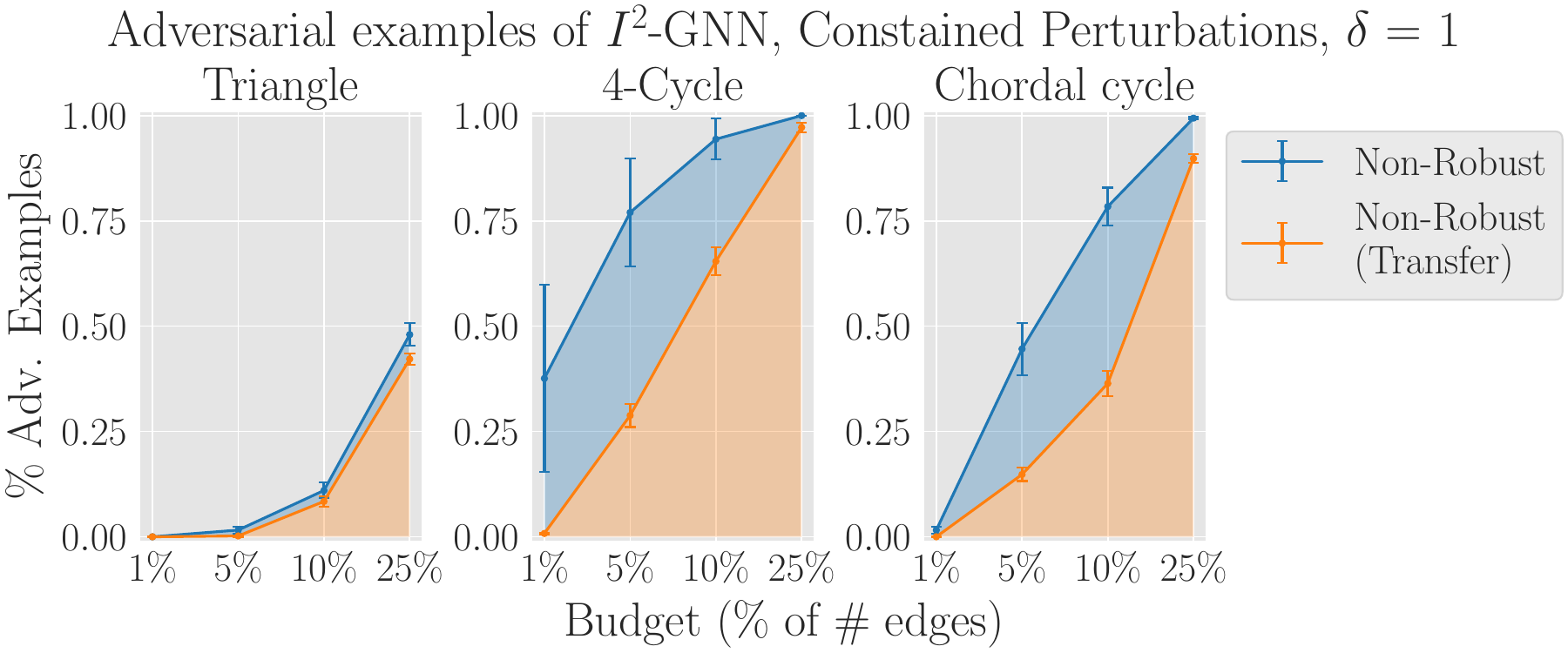}
    \end{subfigure}
    \begin{subfigure}
    \centering
    \includegraphics[width=\columnwidth]{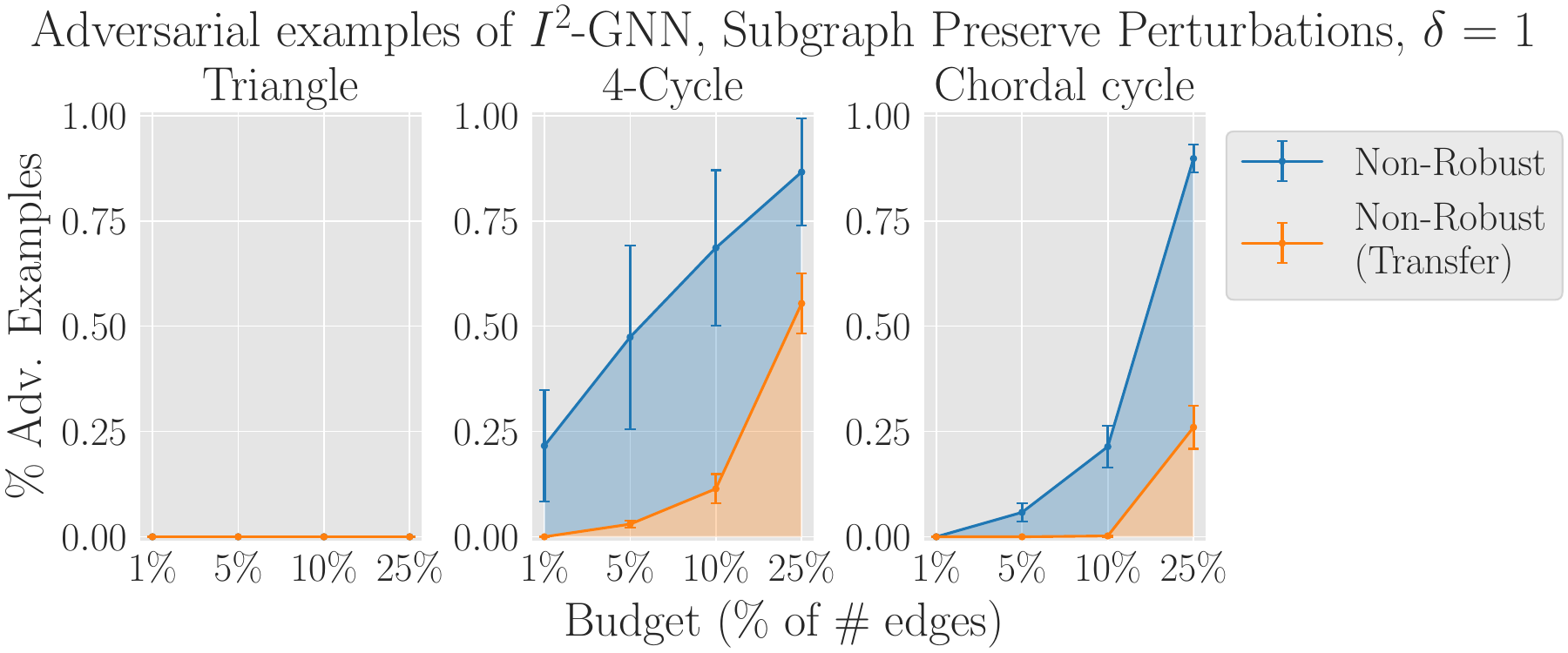}
    \end{subfigure}
    \caption{The plots illustrate in blue the success rate of our subgraph-counting adversarial attacks at finding perturbations that represent adversarial examples according to \cref{def:adv example} constrained and subgraph preserving perturbation spaces. In orange, we represent how effective the adversarial examples are when transferred to the models trained with a different initialization seed.
    The values are the average of the results obtained with 5 different initialization seeds with the relative standard errors.}
    \label{fig:rob plots}
\end{figure}

\textbf{Results.}
The plots in \cref{fig:rob plots} show the percentage of perturbations found by the optimization algorithms that represent a successful adversarial example according to \cref{def:adv example}. 
To condensate the results in a numerical value, we report in \cref{tab:rob exp} the \ac{AUC} of the functions Non-Robust and Non-Robust (Transfer) in \cref{fig:rob plots}. The results are reported as the proportion with respect to the area under the unity function $f(x) = 1$, which represents the worse case where all permutations generate an adversarial example already at $\Delta = 1\%$. 
Interestingly, the results show that we can find several adversarial examples for both architectures. In particular, PPGN is highly unrobust in the subgraph-counting of patterns with four nodes.
However, several adversarial examples can be found also for the triangle count, even though the theoretical expressivity of PPGN claims that it is a family of functions that can count 3-dimensional subgraphs in the sense of \cref{def:subgraph_counting}. Similarly, the more expressive model I$^2$-GNN is fooled on 4-dimensional patterns, in spite of being sufficiently powerful to count them.
This indicates that the empirical expressivity achieved does not match the theoretical expressivity since the models are not able to generalize to subgraph-counting tasks that they should in theory be able to solve. Additionally, in \cref{sec:struct adv} we investigate some structural properties of the adversarial examples.

\begin{table}
\centering
\caption{\ac{AUC} of the functions in \cref{fig:rob plots} normalized by the area under the unity function. The label NR stands for Non-Robust and NR (tr.) for Non-Robust (Transfer).}
\label{tab:rob exp}
\resizebox{\columnwidth}{!}{
    \begin{tabular}{cccccccc}
        \toprule
        Pert. & \multirow{2}{*}{Arch.}& \multicolumn{2}{c}{Trangle} & \multicolumn{2}{c}{4-Cycle} & \multicolumn{2}{c}{Chord. C.}\\
        \cmidrule(lr){3-4}
        \cmidrule(lr){5-6}
        \cmidrule(lr){7-8}
        Space && NR &  NR (tr.) & NR &  NR (tr.) & NR &  NR (tr.) \\
        \midrule
        \midrule
        \multirow{2}{*}{$\pert_\Delta$} & PPGN & 0.18 & 0.10 & 0.95 & 0.87 & 0.95 & 0.86\\& I$^2$-GNN & 0.20 & 0.17 &  0.88 & 0.63 & 0.72 & 0.46\\
        \midrule
        \multirow{2}{*}{$\pert_\Delta^c$} & PPGN & 0.090 & 0.050 & 0.92 & 0.84 & 0.93 & 0.73\\& I$^2$-GNN & 0.0 & 0.0 &  0.76 & 0.40 & 0.41 & 0.097\\ %
        \midrule
        \multirow{2}{*}{$\pert_\Delta^s$} &  PPGN & 0.09 & 0.50 & 0.85 & 0.58 & 0.90 & 0.59\\& I$^2$-GNN & 0.0 & 0.0 &  0.66 & 0.23 & 0.38 & 0.082\\ %
        \bottomrule
    \end{tabular}
}
\end{table}

\subsection{Out-of-Distribution Generalization} \label{sec:exp_ood}
Next, we examine the performance of the same architectures when tested on graphs that are \ac{OOD} with respect to the original training. %
Here, we retrain the last \ac{MLP} prediction layers to  investigate more in detail whether the problem lies in the expressivity of the extracted graph representations or in the last layers that leverage the graph representations for the predictions.
\par
\textbf{Dataset.} For these experiments we use Erdos-Renyi graphs with 10 nodes and edge generation probabilities of 0.3 for the first dataset (d$_1$) and 0.8 for the \ac{OOD} dataset (d$_2$). The size and splits of the datasets are analogous to \cref{sec:exp adv rob}. Therefore, we consider as \ac{OOD} graphs that are generated from a distribution with the same number of nodes but different edge generation probabilities.
\par
\textbf{Experimental Settings.} Firstly, we train the PPGN and I$^2$-GNN architectures on the dataset d$_1$ and test them on both d$_1$ and d$_2$ to investigate the \ac{OOD} generalization performances of the architectures. Additionally, we train the models directly on d$_2$ to have a comparison of the best performances achievable on this dataset. The errors are expressed using the mean absolute error ($\ell_1$) and an extension of it, which is obtained by normalizing by the ground-truth count ($\ell_c$).
\par
\textbf{Results.} \cref{tab:ood} shows the test errors of the aforementioned settings averaged over five different initialization seeds. Here we observe that the models achieve very poor performances on general \ac{OOD} graphs compared to their ideal performances (\ac{OOD} and d$_2$ rows). However, if the model were able to perform subgraph-counting, as theoretically claimed, they should be able to perform this task regardless of the graph distribution. This result matches with \cref{sec:exp adv rob} and shows that the models do not learn to detect the patterns and they rather overfit on the training distribution. However, this behavior could be intrinsic to the models' architecture. The models are designed to extract a vector representation from each input graph, which is then mapped to the prediction through an \ac{MLP}.
Then, the fact that different graph distributions might generate different graph representations leads us to investigate whether the problem is a poor generalization of the map between the graph embedding and the count.
To test this possibility, we retrain on d$_2$ only the final \ac{MLP} of the models previously trained on d$_1$ (row \ac{MLP} in \cref{tab:ood}). While this adjustment is helpful, the errors are consistently one order of magnitude higher than the ones obtained training directly on d$_2$. This indicates that the graph representations do not achieve their theoretic separation power and that the problem does \emph{not} only lie in the last MLP prediction layers.

\begin{table}
\vspace{-0.33cm}
\centering
\caption{Test errors of the \ac{OOD} experiments that investigate the generalization abilities of the architectures. Specifically, d$_i$ represents models trained and tested on the same dataset d$_i$, \ac{OOD} models trained on d$_1$ and tested in d$_2$ and in \ac{MLP} we additionally retrain the final layers on d$_2$.}
\label{tab:ood}
    \resizebox{\columnwidth}{!}{
    \begin{tabular}{cccccccc}
        \toprule
        \multirow{2}{*}{Arch.}& Exp. & \multicolumn{2}{c}{Trangle} & \multicolumn{2}{c}{4-Cycle} & \multicolumn{2}{c}{Chord. C.}\\
        \cmidrule(lr){3-4}
        \cmidrule(lr){5-6}
        \cmidrule(lr){7-8}
        &Setting&$\ell_1$ & $\ell_c$ & $\ell_1$ & $\ell_c$ & $\ell_1$ & $\ell_c$ \\
        \midrule
        \midrule
        \multirow{4}{*}{PPGN} & d$_1$ & 0.0058 & 7.8e-4 & 0.059 & 0.010 & 0.10 & 0.011\\& \ac{OOD}  & 2.98 & 0.041 &  5.40 & 1.17 & 20.0 & 0.25\\
        & d$_2$ & 0.0091 & 1.7e-4 &  0.040 & 0.0050 & 0.12 & 0.0017\\
        & \ac{MLP} & 0.059 & 9.8e-4&  0.29 & 0.0.043 & 1.083 & 0.014\\
        \midrule
        \multirow{4}{*}{I$^2$-GNN} & d$_1$ & 0.0027 & 2.8e-4 & 0.035 & 0.0062 & 0.020 & 0.0023\\
        & \ac{OOD}& 3.25 & 0.044 & 2.16 & 0.45 & 6.75 & 0.084\\
        & d$_2$ & 0.032 & 6.2e-4 &  0.028 & 0.0031 & 0.30 & 0.0042\\
        & \ac{MLP} & 0.20 & 0.0031 &  0.19 & 0.025 & 1.56 & 0.020\\

        \bottomrule
    \end{tabular}
    }
    \vspace{-0.28cm}
\end{table}

%% file: include/7-conclusion.tex
\section{Conclusion}
We proposed a novel approach to assess the empirical expressivity achieved by subgraph-counting \acp{GNN} via adversarial robustness. We show that despite being theoretically capable of counting certain patterns, the models lack generalization as they struggle to correctly predict adversarially perturbed and \ac{OOD} graphs. Therefore, the training algorithms are not able to find weights corresponding to a maximally expressive solution. Extending our study to other related GNNs such as KP-GNN \citep{feng2022how} or to include adversarial training \citep{gosch2023adversarial, geisler_generalization_2022} to steer towards more robust and expressive solutions, are interesting directions for future work. %

\newpage
\section{Acknowledgement}
This paper has been supported by the DAAD programme Konrad Zuse Schools of Excellence in Artificial Intelligence, sponsored by the Federal Ministry of Education and Research.

%% file: include/Appendix.tex
\section{Proof of \cref{thm:ego_pert}}
\label{sec:thm proof}
\begin{proposition}
    Let consider a graph $G$ and a pattern $H$ with $\diam(H) = d$. Then, for every edges $\{i,j\}$ we have that $\ego_d(i)$ and $\ego_d(j)$ contain all the subgraphs $G_S \subset G$ such that $G_S \simeq H$ and $i,j \in V_S$.
    \label{thm:ego_pert2}
\end{proposition}
\emph{Proof.} Let's consider any subgraph $G_S = (V_S, E_S)$ of the graph $G = (V,E)$ such that $G_S \simeq H$. Firstly we show that $\ego_d(i) = (V_e, E_e)$ contains $G_S$, that is equivalent to showing that $V_S \subseteq V_e$ and $E_S \subseteq E_e$. By construction, we have that $\diam(H) = \diam(G_S) = d$, which implies that for every node $ l \in V_S$ the shortest path connecting $i$ and $l$ has at most length $d$. Therefore, by definition of egonet, we have that $V_S \subseteq V_e$
Moreover, we know by construction that $E_S \subseteq E$ and can write the following series of set inclusions:
\[E_S  \subseteq V_S \times V_S \subseteq V_e \times V_e.\]
All in all, we have that:
\[E_S \subseteq E \cap (V_e \times V_e) = E_e.\]
Showing that $\ego_d(j)$ contains $V_S$ is analogous.
\par
We can also easily show that for any $\ego_l(i)$ (or $\ego_l(j)$) with $l > d$ \cref{thm:ego_pert2} does not hold. In fact, it is sufficient to consider $i$ such that there exists a pair of nodes $\{i,l\}$ where the shortest path connecting the two nodes has length $d$. Therefore, by construction $l \notin V_e$ and \cref{thm:ego_pert2} does not hold. 
\par
\section{Counting algorithm}
\label{sec:counting alg}
\begin{figure}[ht]
    \centering
    \begin{tabular}{cccc}
        \centering
         {\includegraphics[width = 0.15\textwidth]{pictures/triangle.png}}&
         {\includegraphics[width = 0.15\textwidth]{pictures/2-path.png}}&
         {\includegraphics[width = 0.15\textwidth]{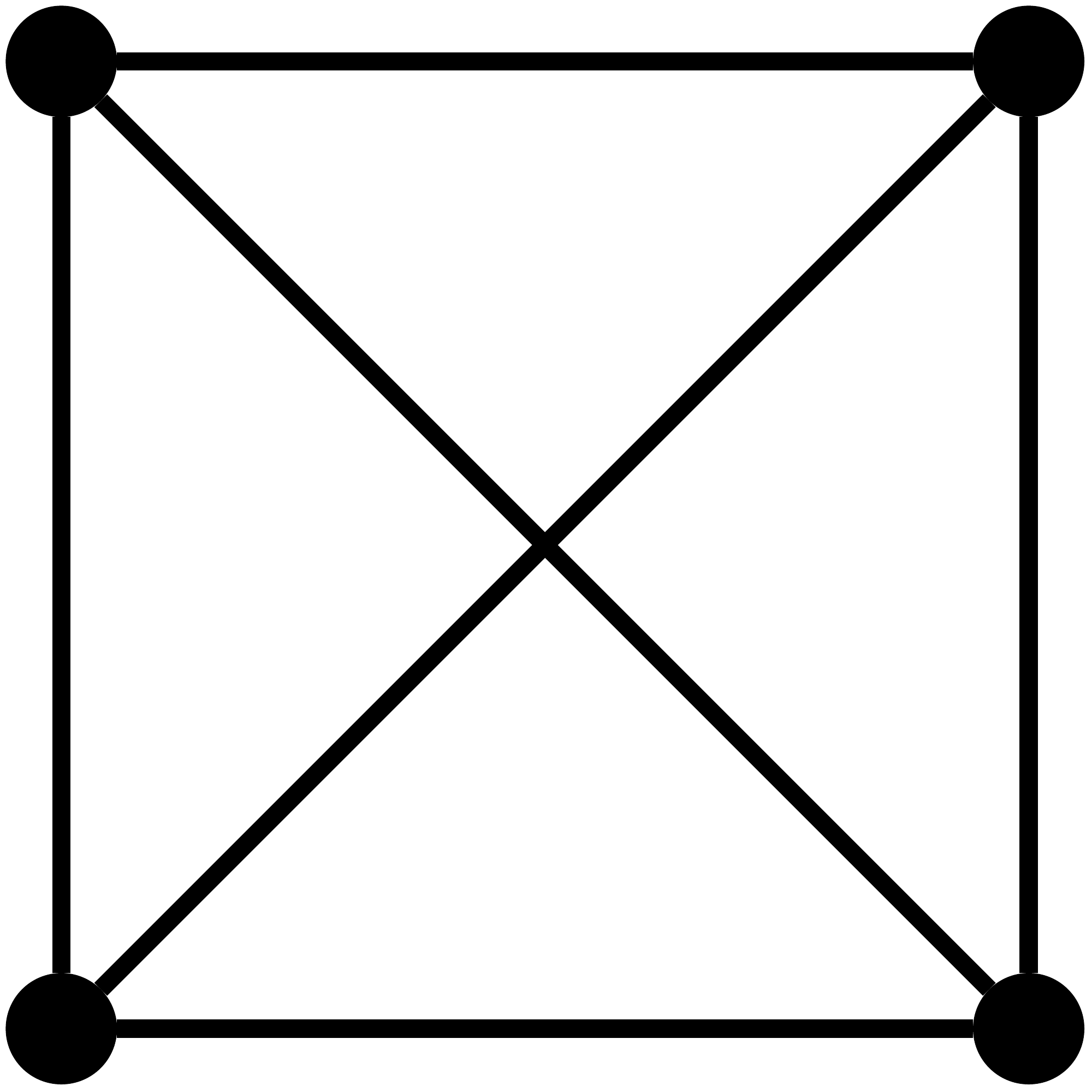}}&
         {\includegraphics[width = 0.15\textwidth]{pictures/chord_ccycle.png}}\\
         Triangle $(c_1^{(3)})$ & 2-Path $(c_2^{(3)})$&4-Clique &Chordal cycle\\
         \vspace{2mm}\\
         {\includegraphics[width = 0.15\textwidth]{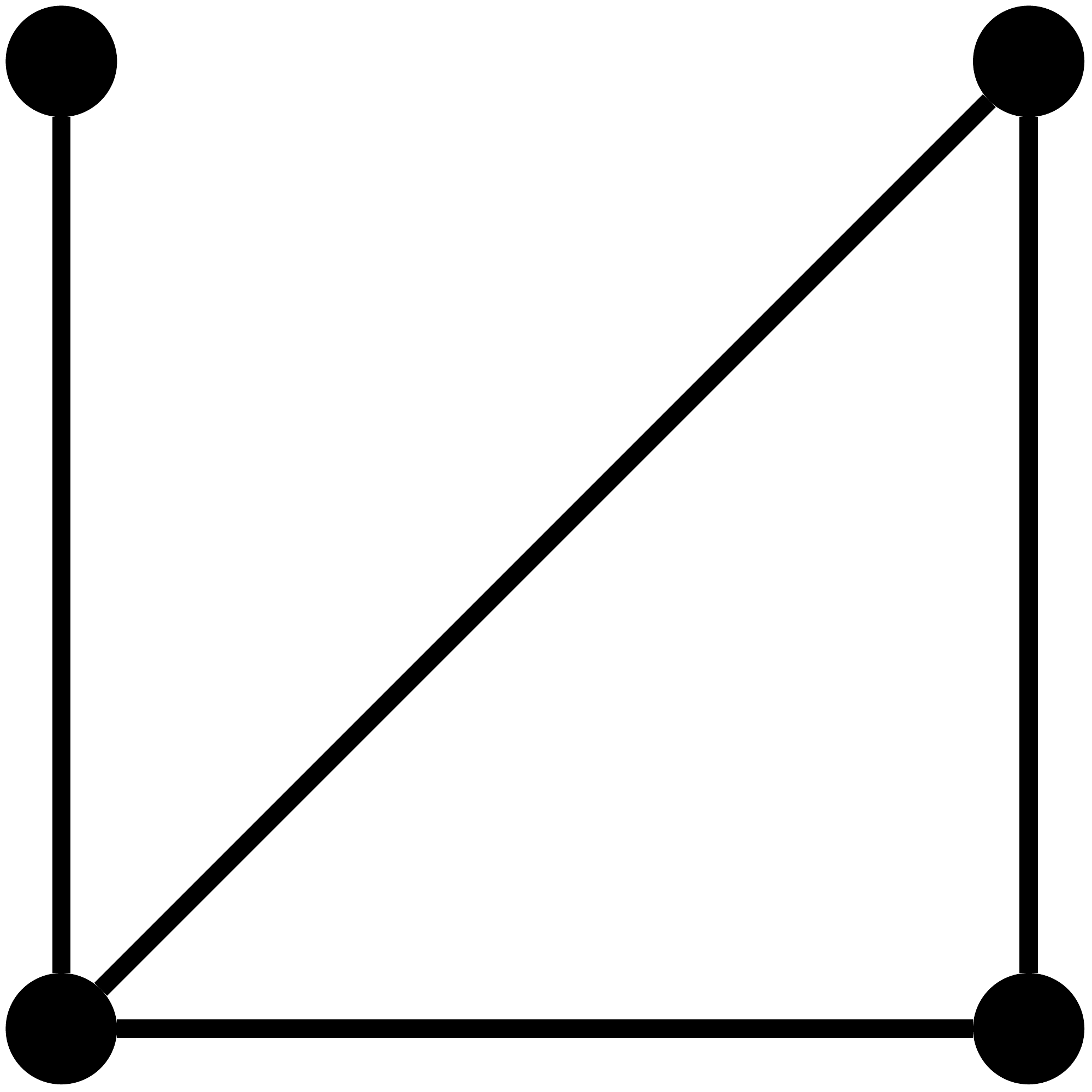}}&
         {\includegraphics[width = 0.15\textwidth]{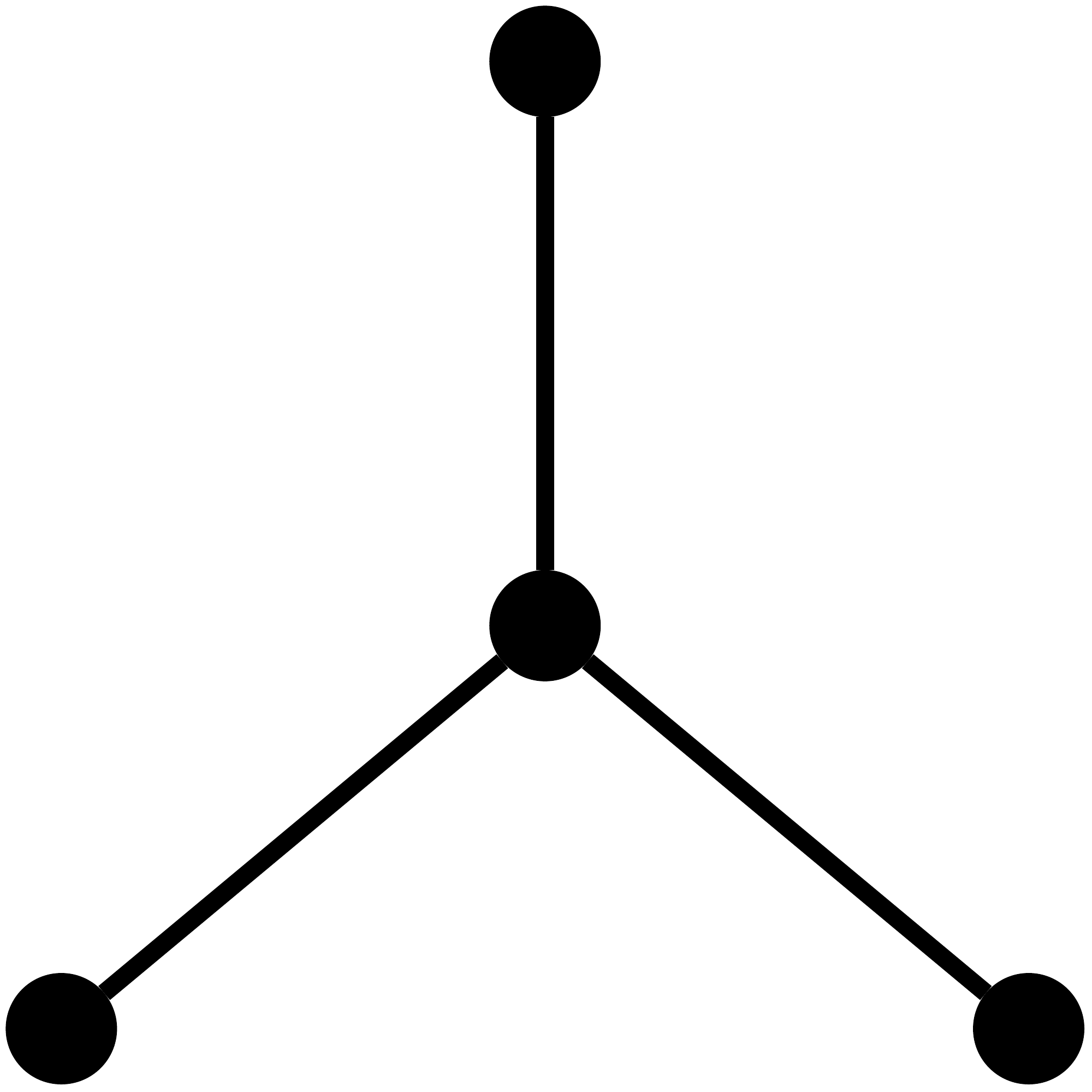}}&
         {\includegraphics[width = 0.15\textwidth]{pictures/4-cylce.png}}& \includegraphics[width=0.15\textwidth]{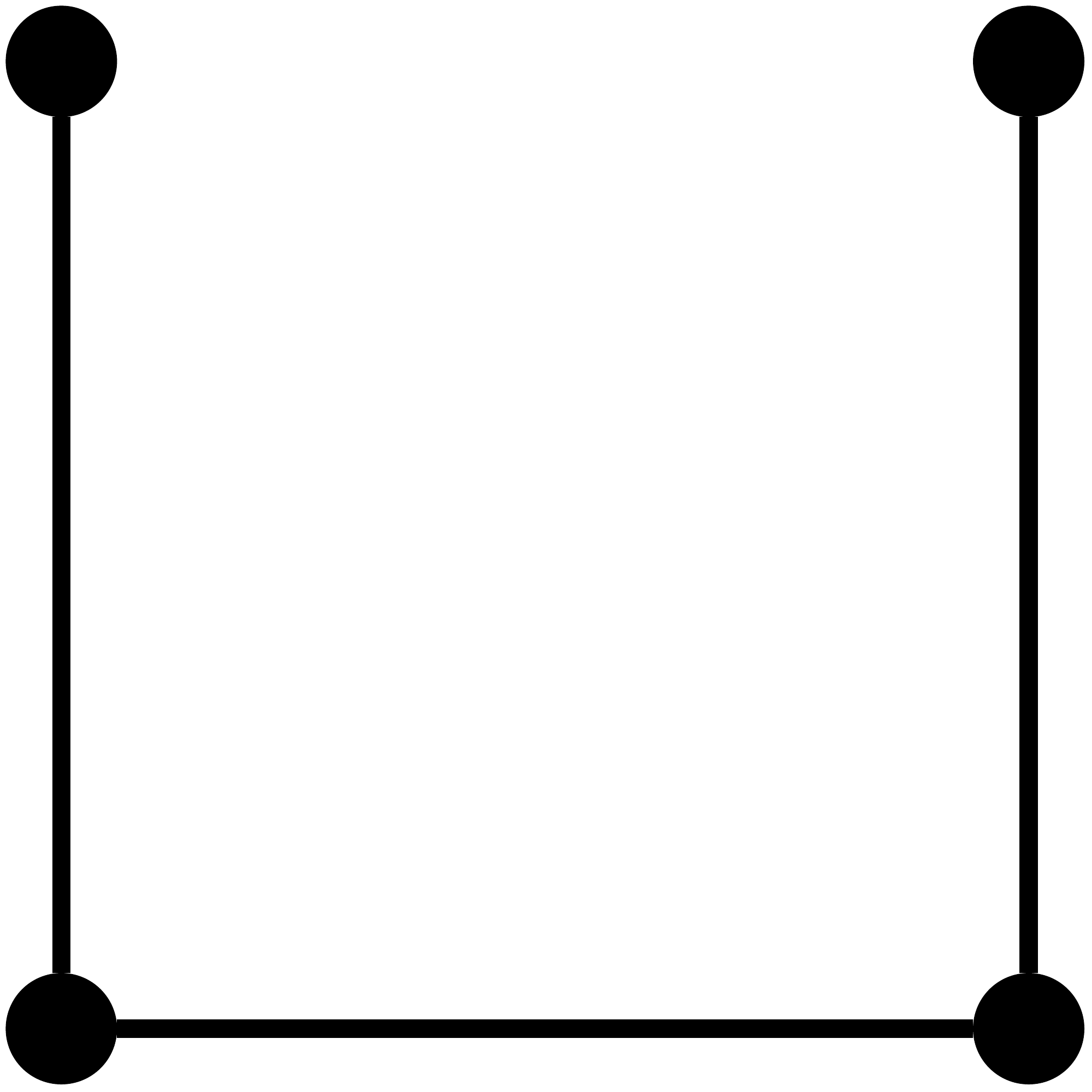}\\
         Tailed triangle& 3-Star &4-Cycle&3-Path
    \end{tabular}

    \caption{Connected subgraphs with 3 or 4 nodes. In brackets the name with which we represent the count variables in \cref{alg:3-dim}}
    \label{fig:graphlets}
\end{figure}

To generate synthetic datasets and to perform adversarial attacks (\cref{sec:adv att}) we need an algorithm that is capable of counting subgraphs. However, subgraph counting is a highly complex procedure, and naive  approaches could scale to a prohibitive amount of computations. As a matter of fact, only enumerating all the possible $k$-dimensional subgraph  requires $\binom{n}{k}$ iterations, which can become intractable for sufficiently large graphs.
Since the computation of groud-truth will be extensively used by adversarial attacks, finding an efficient algorithm to count subgraphs is in our interest to reduce the overall computational complexity of the attacks. \citep{shervashidze_ecient_nodate} proposed a method to count induced graphlets of size 3, 4, and  5 with a computational complexity of $\mathcal{O}(Nd^{k-1})$,  where $d$ is the maximum node degree and $k$ is the size of the graphlets, which is linear in the number of nodes and is especially efficient sparser graphs. In particular, in this work we will consider only connected subgraphs of 3 and 4 nodes to have a broad overview of the performances on patterns with different structural properties (\cref{fig:graphlets}).

The core idea of the algorithm is to extract connected paths of length $k-1$ and a single node connected to the path, then this set of nodes is assigned to a substructure using handcrafted conditions based on the additional connections between  nodes. For example, for a 2-path $\{v_1, v_2, v_3\}$ and an additional node $v_4$ that is connected only to the head and the tail of the path, one can conclude that the four nodes generate a 4-Cycle.  
\par
The algorithm that counts 3-dimensional graphlets (\cref{alg:3-dim}) starts by considering every edge $\{v_1, v_2\}$ in the graph, which has $\mathcal{O}(Nd)$ complexity. Next, one can count the subgraphs with specific computations on the neighboring sets of $v_1$ and $v_2$, in particular, the nodes in both neighborhoods will constitute a triangle, and the ones belonging only to one neighborhood will constitute a 2-path. This last step has $\mathcal{O}(d)$ complexity because every neighborhood contains at most $d$ nodes. Finally, every subgraph is detected more than once as every pattern can be identified starting with any of its edges, then every variable is normalized by dividing by the number of edges of the pattern it represents. 
\par
The 4-dimensional version has an analogous working principle adapted to the 4 four nodes case. 
As before, the algorithm iterates over every edge in the graph, then all the nodes $v_3$ adjacent to this edge ($\mathcal{O}(d)$) are considered and separated into 3 disjoint groups based on the connections with $\{v_1, v_2\}$ (connected with both nodes, only the first one or only the second one). 
At this point, similarly to before, the patterns can be identified with specific computations of the neighborhoods of $v_1, v_2, v_3$. For example, if $v_3$ is only  in the neighborhood of $v_1$ and $v_4$ connected to all the previous nodes, then $\{v_1, \dots, v_4\}$ is a tailed triangle. In the last step, the variables need to be normalized to take into account multiple detections of the same subgraph. 
By construction, once the triplet $\{v_1, v_2, v_3\}$ is set, the algorithm can detect a subgraph containing this triplet only once, hence normalization constant is the number of times the algorithm obtains a triplet included in a subgraph. In particular, triangles can be obtained in three different ways, starting from each edge, and, for the same reason, 2-paths can be obtained in two ways. 
All in all, the normalization constant for a general pattern is $2p + 3t$, where $p$ is the number of 2-paths in the pattern and $t$ is the number of triangles.
Moreover, these algorithms can not only count the induced subgraphs, but also identify them by returning the set of nodes $N' \subset N$ that generates them, which will be essential for subgraph preserving attacks (\cref{sec:adv att}).

\begin{algorithm}[h]
\caption{Count 3-dimensional induced subgraphs in a graph $G=(N, E)$}
\label{alg:3-dim}
\begin{algorithmic}
        \STATE $c_i^{(3)} \leftarrow 0, \ i \in \{1, 2\}$ \hfill\COMMENT{Initialize the variables}
        \FOR{$\{v_1, v_2\} \in E$}
            \STATE $c_1^{(3)} \leftarrow c_1^{(3)} + \big|\mathcal{N}(v_1) \cap \mathcal{N}(v_2)\big|$ \hfill\COMMENT{Update the substructure counts}
            \vspace{2mm}
            \STATE $c_2^{(3)} \leftarrow c_2^{(3)} + \big|\mathcal{N}(v_1) / \big(\mathcal{N}(v_2) \cup \{v_2\}\big)\big|$
            \STATE$c_2^{(3)} \leftarrow c_2^{(3)} + \big|\mathcal{N}(v_2) / \big(\mathcal{N}(v_1) \cup \{v_1\}\big)\big|$
        \ENDFOR
        \STATE $c_1^{(3)}(G) \leftarrow c_1^{(3)}(G)/3$ \hfill\COMMENT{Normalize the variables }
        \STATE $c_2^{(3)}(G) \leftarrow c_2^{(3)}(G)/2$
        \STATE {\bfseries Return:} $c_i^{(3)}, \ i \in \{1, 2\}$
\end{algorithmic}
\end{algorithm}

\section{Additional Results}
\label{sec:add res}
In this section, we extend the experiments of \cref{sec:exp} to all the connected patterns with three and four nodes \cref{fig:graphlets}. In particular, in \cref{sec:adv rob appendix} we extend the adversarial robustness experiments, in \cref{sec:struct adv} we compare some structural properties of the adversarial examples with the clean graphs, and in \cref{sec:ood appendix} we present the complete results of the \ac{OOD} experiments. 
\subsection{Adversarial Robustness}
\label{sec:adv rob appendix}
\begin{figure}[!h]
    \centering
    \includegraphics[width=\textwidth]{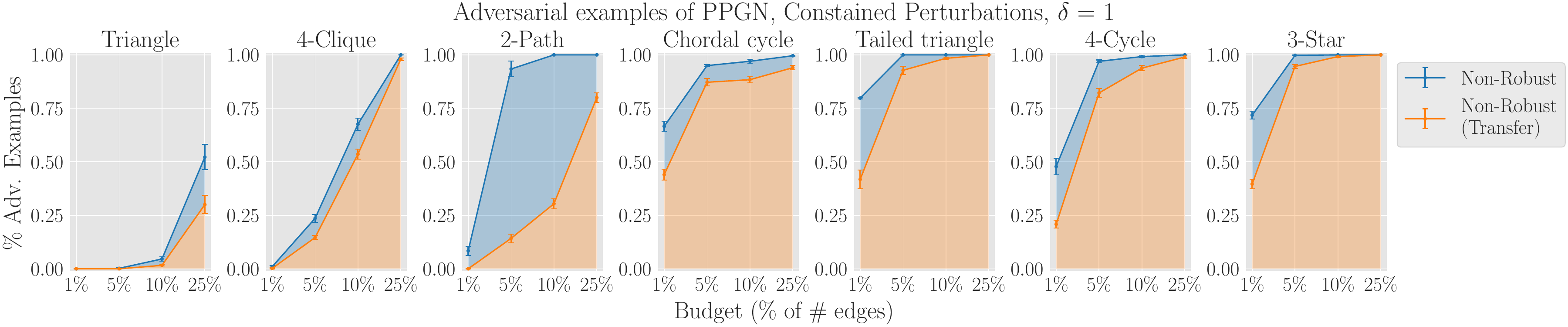}
    \includegraphics[width=\textwidth]{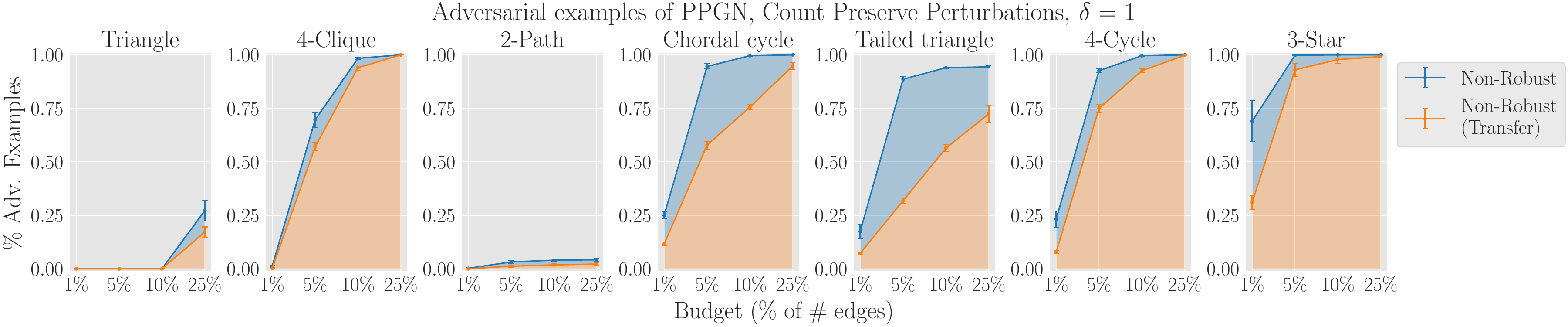}
    \includegraphics[width=\textwidth]{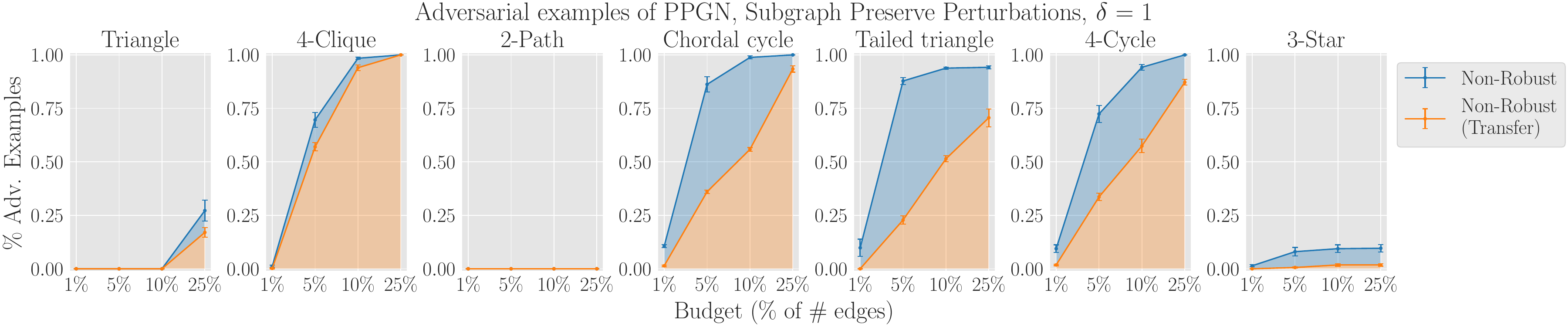}
    \caption{The plots illustrate in blue the success rate of our subgraph-counting adversarial attacks at finding perturbations that represent adversarial examples according to \cref{def:adv example} in the three perturbation spaces we defined for PPGN. In orange, we represent how effective the adversarial examples are when transferred to the models trained with a different initialization seed.
    The values are the average of the results obtained with 5 different initialization seeds with the relative standard errors. }
    \label{fig:res PPGN}
\end{figure}

\begin{figure}[!h]
    \centering
    \includegraphics[width=\textwidth]{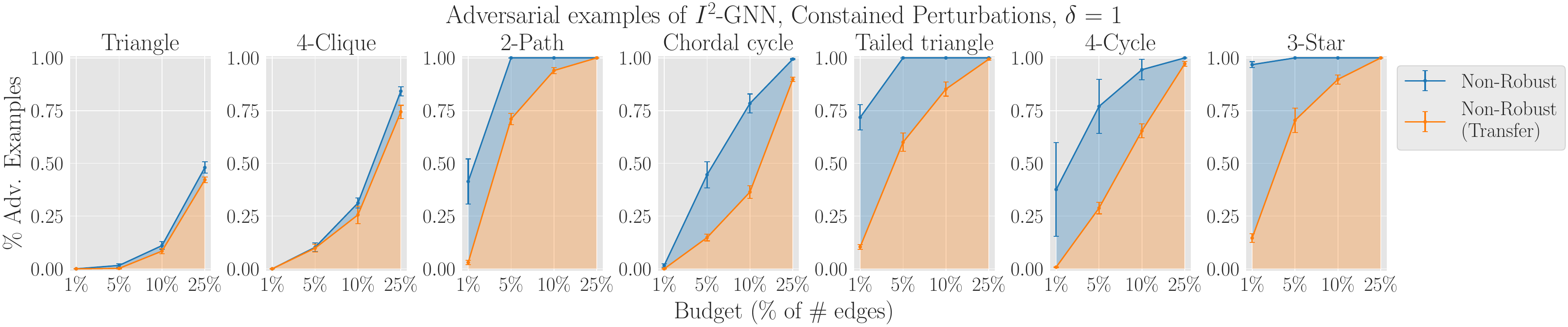}
    \includegraphics[width=\textwidth]{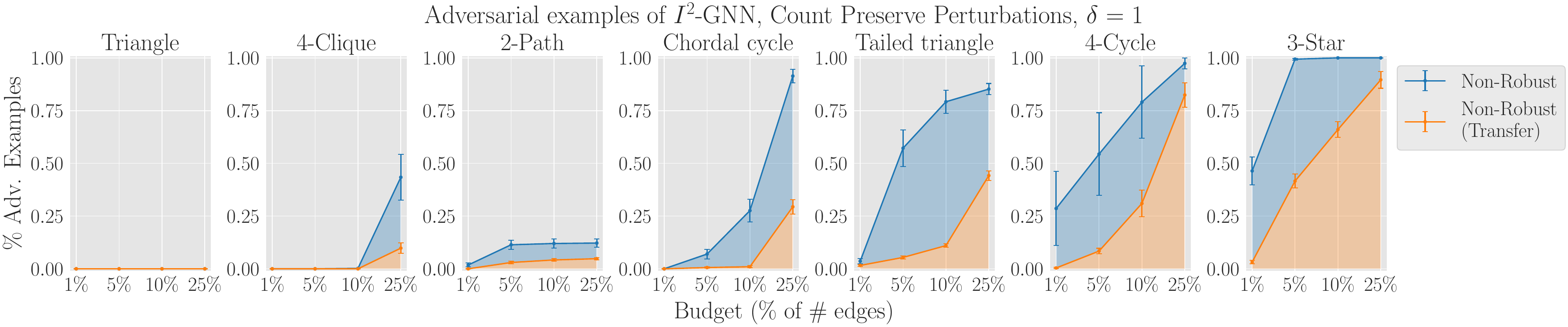}
    \includegraphics[width=\textwidth]{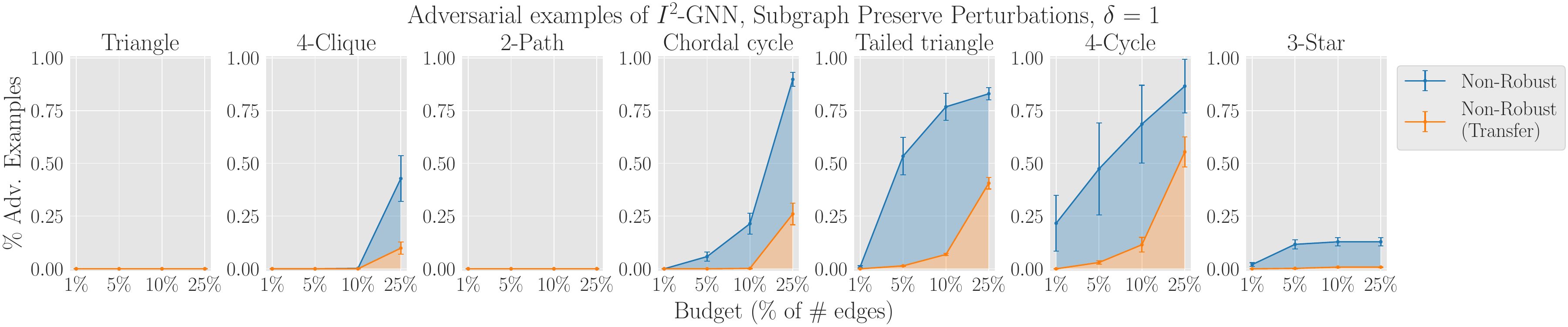}
    \caption{The plots illustrate in blue the success rate of our subgraph-counting adversarial attacks at finding perturbations that represent adversarial examples according to \cref{def:adv example} in the three perturbation spaces we defined for I$^2$-GNN. In orange, we represent how effective the adversarial examples are when transferred to the models trained with a different initialization seed.
    The values are the average of the results obtained with 5 different initialization seeds with the relative standard errors. }
    \label{fig:res I2GNN}
\end{figure}

In \cref{fig:res PPGN} we present the complete adversarial robustness results of PPGN on all the 3- and 4-dimensional patterns on all the 3 perturbations paces ($\pert_\Delta, \pert_\Delta^c$ and $\pert_\Delta^s$). In \cref{fig:res I2GNN} we present the analogous results for I$^2$-GNN. The dataset and experimental settings are identical to \cref{sec:exp adv rob}. Additionally, in \cref{tab:loss} we include the training errors of the models with respect to the MAE ($\ell_1$) and the MAE normalized by the ground-truth count ($\ell_{c}$). Specifically, we present the average error over five models trained using different initialization seeds and the standard errors. Note $\ell_1$ error of 3-Path is considerably higher than 0.5, which makes the search of correctly predicted test graphs almost impossible. For this reason, we excluded the 3-Path pattern from the adversarial robustness experiments.
On the new patterns, the results remain aligned to the discussion in \cref{sec: adv rob} except for 2-Path in $\pert_\Delta^c, \pert_\Delta^s$, and 3-Star in $\pert_\Delta^s$. However, these specific cases are caused by the fact that the perturbation spaces contain almost no elements (see bold elements in \cref{tab:num pert}).

\begin{table}[t]
\caption{Test error of the architectures GIN, PPGN, and I$^2$-GNN on the Stochastic Block Model dataset with respect to $\ell_1$ and $\ell_c$ metrics for different subgraph-counting tasks. The results are expressed as the average over the 5 different weight initialization $\pm$ the standard error.}
\centering
\label{tab:loss}
    \resizebox{\textwidth}{!}{
    \begin{tabular}{cccccccccc}
        \toprule
        Arch. & Loss & Triangle & 2-Path & 4-Clique & Chord. C. & Tailed Tr. & 3-Star & 4-Cycle & 3-Path\\

        \midrule
        \midrule
        
        \multirow{2}{*}{GIN} & $\ell_1$ & 2.6 $\pm$ 0.017 & 7.6 $\pm$ 0.062 & 0.81 $\pm$ 0.0041 & 4.4 $\pm$ 0.0087 & 15.0 $\pm$ 0.14 & 21.0 $\pm$ 0.076 & 6.3 $\pm$ 0.012 & 40.0 $\pm$ 0.15  \\
        
        & $\ell_c$ & 0.13 $\pm$ 6.2e-4 & 0.03 $\pm$ 2.3e-4 & 0.34 $\pm$ 0.0015 & 0.27 $\pm$ 0.0019 & 0.098 $\pm$ 7.8e-4 & 0.087 $\pm$ 4.0e-4 & 0.18 $\pm$ 0.0015 & 0.053 $\pm$ 1.5e-4  \\
        
        \midrule
        
        \multirow{2}{*}{PPGN} & $\ell_1$ & 0.014 $\pm$ 5.5e-4 & 0.15 $\pm$ 0.0098 & 0.038 $\pm$ 0.001 & 0.38 $\pm$ 0.0069 & 1.1 $\pm$ 0.065 & 0.81 $\pm$ 0.075 & 0.23 $\pm$ 0.0083 & 2.7 $\pm$ 0.12  \\
        
        & $\ell_c$ & 6.9e-4 $\pm$ 2.2e-5 & 6.0e-4 $\pm$ 4.0e-5 & 0.013 $\pm$ 6.5e-4 & 0.017 $\pm$ 4.7e-4 & 0.0061 $\pm$ 4.0e-4 & 0.0033 $\pm$ 3.0e-4 & 0.0061 $\pm$ 2.0e-4 & 0.0035 $\pm$ 1.7e-4  \\

        \midrule
        
        \multirow{2}{*}{I$^2$-GNN} & $\ell_1$ & 0.0069 $\pm$ 4.4e-4 & 0.26 $\pm$ 0.025 & 0.0031 $\pm$ 2.9e-4 & 0.058 $\pm$ 0.005 & 0.39 $\pm$ 0.022 & 0.6 $\pm$ 0.042 & 0.25 $\pm$ 0.093 & 1.5 $\pm$ 0.31  \\
        
        & $\ell_c$ & 3.8e-4 $\pm$ 3.0e-5 & 0.001 $\pm$ 9.9e-5 & 6.7e-4 $\pm$ 7.0e-5 & 0.0028 $\pm$ 2.2e-4 & 0.0022 $\pm$ 1.4e-4 & 0.0024 $\pm$ 1.7e-4 & 0.0068 $\pm$ 0.0026 & 0.0021 $\pm$ 4.0e-4  \\
        
        \bottomrule
    \end{tabular}
    }
\end{table}

\begin{table}[h]
\caption{Average size of the preserving perturbations with budget 1. The values represent an estimation of the number of alternatives that each step of the search algorithm has.}
\centering
\label{tab:num pert}
    \begin{tabular}{cccccccccc}
        \toprule
        & Trangle & 2-Path & 4-Clique & Chord. C. & Tailed Tr. & 3-Star &  4-Cycle & 3-Path\\
    
        \midrule
        \midrule
        $|\pert_1^c(G)|$ & 205.32 & \textbf{1.69} & 410.93 & 244.04 & 27.94 & 12.24 & 88.98 & 1.64\\
        $|\pert_1^s(G)|$ & 205.32 & \textbf{0.02} & 410.93 & 241.82 & 27.62 & \textbf{1.6} & 74.34 & 0.62\\

        \bottomrule
    \end{tabular}
\end{table}

\subsection{Structural Properties of Adversarial Examples}
\label{sec:struct adv}
\begin{figure}
    \centering
    \includegraphics[width=\textwidth]{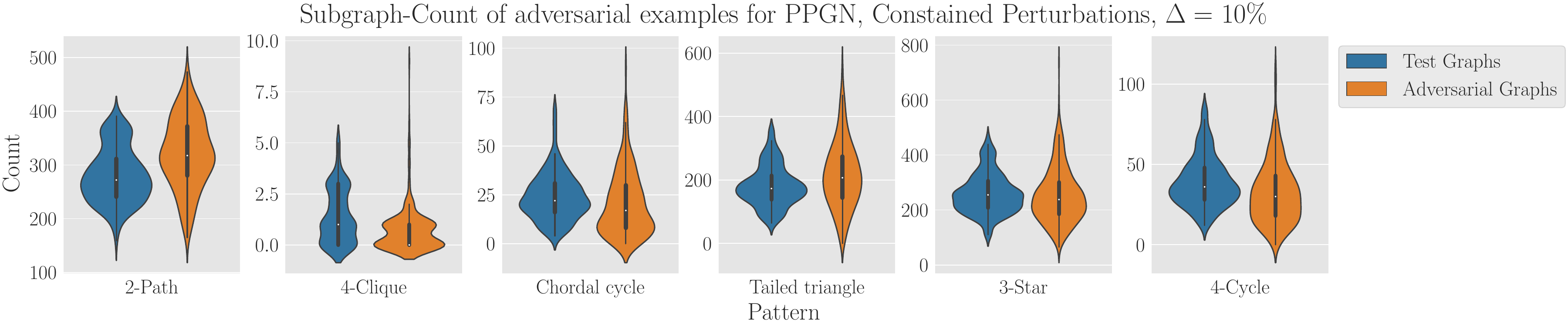}
    \par
    \medskip
    \includegraphics[width=\textwidth]{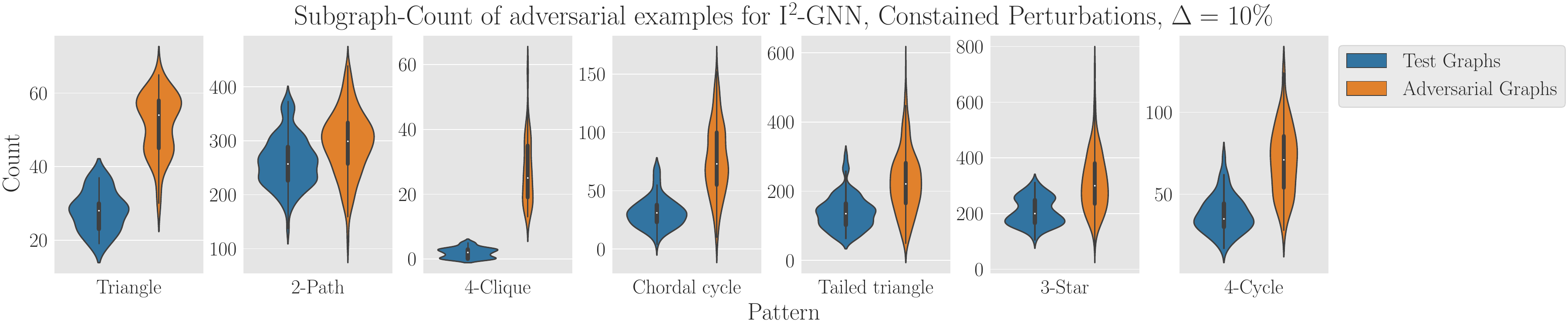}
    \caption{Distribution of the \textbf{subgraph counts} of the transferring adversarial examples (in orange) and their relative clean graphs (in blue). In particular, the adversarial examples are generated from the perturbation space $\pert_{10\%}$, and we present only the patterns having at least 5\% of successful attacks.}
    \label{fig:adv subgraph count}
\end{figure}

\begin{figure}
    \centering
    \includegraphics[width=\textwidth]{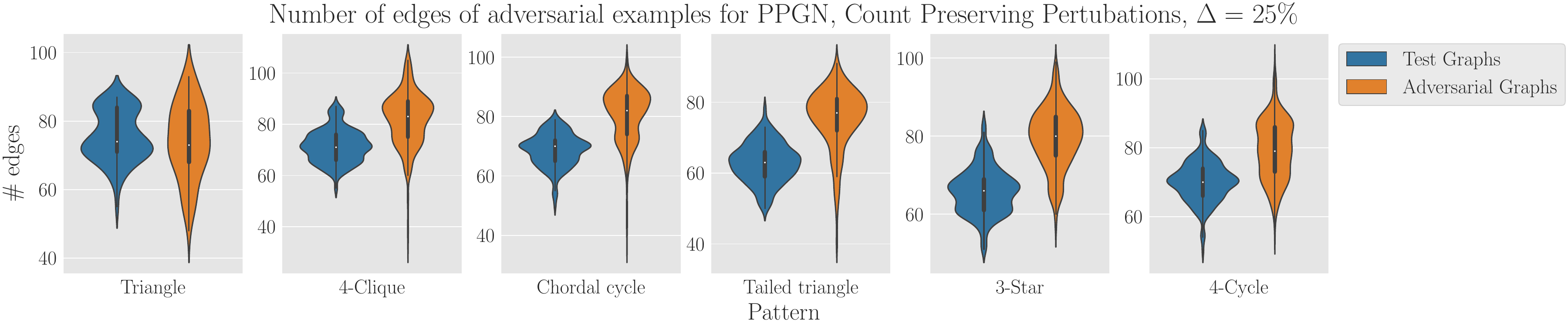}
    \par
    \medskip
    \includegraphics[width=\textwidth]{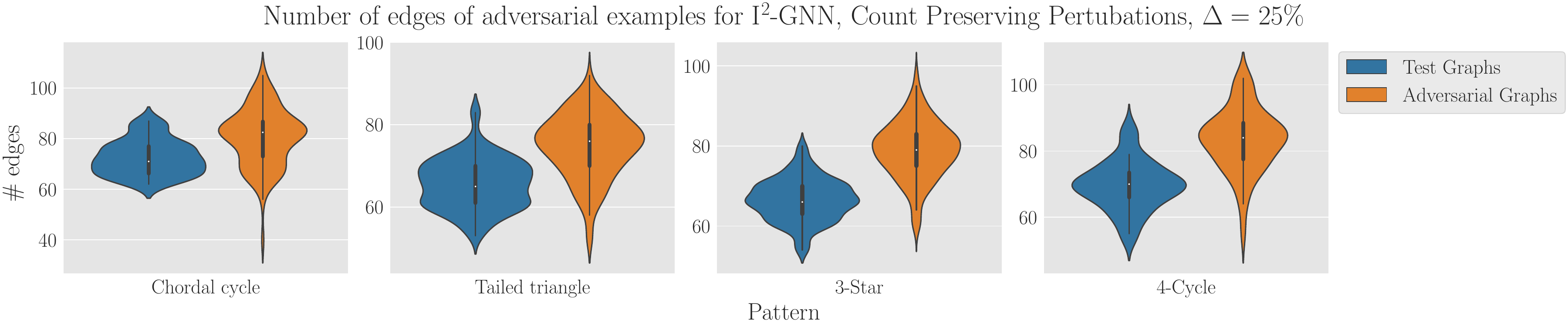}
    \caption{Distribution of the \textbf{number of edges} of the transferring adversarial examples (in orange) and their relative clean graphs (in blue). In particular, the adversarial examples are generated from the perturbation space $\pert^c_{25\%}$, and we present only the patterns having at least 5\% of successful attacks.}
    \label{fig:adv edges}
\end{figure}

The previous experiments identify the adversarial examples for the two more expressive architectures, then we can analyze them to get a few insights about the reasons why the architectures are not robust.
In \cref{fig:adv subgraph count} we compare the distribution of the subgraph counts of the test graphs and the adversarial graphs generated from the constrained perturbation space. This experiment is designed to explore whether the adversarial examples generation exploits the fact that this perturbation space allows changing significantly the semantic meaning of the graphs. 
Concretely, for each pattern, we extract the successful transferring adversarial examples, i.e. perturbations that fool all five trained models, with budget $\Delta = 10 \%$, and compare their subgraph-count distribution to the distribution of the corresponding clean graphs.
We specifically choose to consider exclusively the adversarial examples that transfer to the other models because they intuitively represent the failure modes that affect the architectures in general.
Moreover, we only consider  patterns where at least 5\% of attacks have produced a successful adversarial example.
This makes the representations more reliable since the distributions are estimated over at least 25 samples \footnote{We run 500 attacks, 100 for each one of the 5 models}.
To plot the two distributions we use a \emph{violinplot}, which, differently from a box plot, gives also a visual representation of probability density.
From the plots, we notice a variation in the distribution for several patterns. In particular, all the adversarial distributions have heavier tails, which means that outliers counts are more frequent, and in some cases also the whole distribution is shifted.
This result suggests that altering the ground truth count is a failure mode for both architectures and justifies the introduction of more restrictive perturbation spaces which preserve the subgraph count. 
However, the adversarial attacks are capable of generating adversarial examples also for the count and subgraph preserving spaces, which encourages us to further investigate additional failure modes.
In \cref{fig:adv edges} we continue with this analysis by comparing  the distribution of the number of edges of the adversarial graphs and the clean graphs. 
The plots specifically show the distributions of the transferring adversarial examples generated from the count preserving perturbation space with budget $\Delta = 25\%$ and the corresponding test graphs. 
Here we find that the adversarial graphs tend to have more edges they seem to belong to a different graph distribution.

\newpage
Additionally, we investigate the statistical significance of the aforementioned distributional shifts.
To do so, we compute the p-values of the t-tests (with no equal population variance assumption) for the null hypothesis that the distributions of the adversarial and clean graphs have the same expected value.
\cref{tab:count pval} shows the p-values for the tests for the subgraph count distributions of adversarial examples extracted from the constrained perturbation space and the clean graphs. Instead, \cref{tab:edges pval} shows the analogous result for the distribution of the number of edges and for adversarial graphs searched in the count-preserving perturbation space.
Moreover, as before we only consider the subgraphs and budget from which at least 5\% of the attacks are successful, for all the other cases we omit the p-value.
All in all, the results show that from budget 10\% we can affirm for almost all the distribution shift is statistically significant.
\begin{table}
    \centering
    \caption{P-values that represent the statistical significance of the shift between the \textbf{subgraph count distributions} of the transferring adversarial graph extracted from $\pert_{\Delta}$ and their relative clean graphs. In particular, we report the p-values of a t-test for the null hypothesis that the expected values of the distributions are the same. The p-values of  samples with less than 25 elements are omitted.}
    \begin{tabular}{ccccccccc}
        \toprule
         Arch. & Budget & Triangle & 2-Path & 4-Clique & Chord. C. & Tailed Tr. & 3-Star & 4-Cycle\\
         \midrule
         \midrule
        \multirow{4}{*}{PPGN}& 1$\%$ &  -  &  -  &  -  & 0.4 & 0.0095 & 0.63 & 0.87\\
& 5$\%$ &  -  & 0.0067 & 0.11 & 0.64 & 7.0e-11 & 0.25 & 0.32\\
& 10$\%$ &  -  & 8.4e-10 & 3.0e-11 & 9.7e-5 & 2.6e-9 & 0.19 & 2.8e-9\\
& 25$\%$ & 2.2e-6 & 4.6e-10 & 3.0e-16 & 1.6e-28 & 0.062 & 1.0e-16 & 1.3e-116\\
\midrule
        \multirow{4}{*}{I$^2$-GNN}& 1$\%$ &  -  &  -  &  -  &  -  & 0.04 & 0.15 &  - \\
& 5$\%$ &  -  & 8.2e-6 & 5.7e-38 & 9.5e-17 & 1.9e-22 & 1.5e-22 & 1.6e-10\\
& 10$\%$ & 2.1e-24 & 1.1e-19 & 7.9e-59 & 1.9e-45 & 2.7e-51 & 2.1e-57 & 2.0e-33\\
& 25$\%$ & 2.2e-103 & 1.1e-80 & 1.5e-102 & 8.5e-112 & 3.5e-86 & 1.5e-78 & 7.9e-105\\
         \bottomrule
    \end{tabular}
    
    \label{tab:count pval}
\end{table}

\begin{table}
    \centering
    \caption{P-values that represent the statistical significance of the shift between the \textbf{distributions of the number of edges} of the transferring adversarial graph extracted from $\pert^c_{\Delta}$ and their relative clean graphs. In particular, we report the p-values of a t-test for the null hypothesis that the expected values of the distributions are the same. The p-values of  samples with less than 25 elements are omitted.}
    \begin{tabular}{ccccccccc}
        \toprule
         Arch. & Budget & Triangle & 2-Path & 4-Clique & Chord. C. & Tailed Tr. & 3-Star & 4-Cycle\\
         \midrule
         \midrule
        \multirow{4}{*}{PPGN}& 1$\%$ & - & - & - & 5.2e-5 & 0.49 & 0.89 & 0.0073\\
& 5$\%$ & - & - & 1.1e-4 & 0.53 & 0.0017 & 7.3e-12 & 1.7e-5\\
& 10$\%$ & - & - & 0.0069 & 1.2e-11 & 1.8e-18 & 5.0e-37 & 4.1e-4\\
& 25$\%$ & 0.51 & - & 1.3e-67 & 5.2e-99 & 2.3e-83 & 7.4e-147 & 6.8e-75\\

\midrule
        \multirow{4}{*}{I$^2$-GNN}& 1$\%$ & - & - & - & - & - & 0.69 & -\\
& 5$\%$ & - & - & - & - & 0.32 & 7.9e-6 & -\\
& 10$\%$ & - & - & - & - & 0.029 & 1.1e-20 & -\\
& 25$\%$ & - & - & - & 2.7e-8 & 1.6e-26 & 3.8e-110 & 1.8e-19\\
\bottomrule
    \end{tabular}

    \label{tab:edges pval}
\end{table}

\subsection{Out-of-Distribution Generalization}
\label{sec:ood appendix}
Similarly, we present also the \ac{OOD} experiments results for all the patterns in \cref{tab:ood extended}. Also in this case the discussion of the results in \cref{sec:exp_ood} complies also with the results on the new patterns.

\begin{table}[h]
\caption{Test errors of the \ac{OOD} experiments that investigate the generalization abilities of the architectures. Specifically, d$_i$ represents models trained and tested on the same dataset d$_i$, \ac{OOD} models trained on d$_1$ and tested in d$_2$ and in \ac{MLP} we additionally retrain the final layers on d$_2$.}
\centering
\label{tab:ood extended}
    \resizebox{\textwidth}{!}{
    \begin{tabular}{cccccccccccccccccc}
        \toprule
        \multirow{2}{*}{Arch.}& Exp. & \multicolumn{2}{c}{Trangle}
        & \multicolumn{2}{c}{2-Path} & \multicolumn{2}{c}{4-Clique} & \multicolumn{2}{c}{Chord. C.} & \multicolumn{2}{c}{Tailed Tr.} & \multicolumn{2}{c}{3-Star} & \multicolumn{2}{c}{4-Cycle} & \multicolumn{2}{c}{3-Path}\\
        \cmidrule(lr){3-4}
        \cmidrule(lr){5-6}
        \cmidrule(lr){7-8}
        \cmidrule(lr){9-10}
        \cmidrule(lr){11-12}
        \cmidrule(lr){13-14}
        \cmidrule(lr){15-16}
        \cmidrule(lr){17-18}
        & Setting & $\ell_1$ & $\ell_c$ & $\ell_1$ & $\ell_c$ & $\ell_1$ & $\ell_c$ & $\ell_1$ & $\ell_c$ & $\ell_1$ & $\ell_c$ & $\ell_1$ & $\ell_c$ & $\ell_1$ & $\ell_c$ & $\ell_1$ & $\ell_c$ \\
        \midrule
        \midrule

        \multirow{4}{*}{PPGN} & d$_1$ & 0.0058 & 7.8e-4 & 0.016 & 4.7e-1 & 0.0098 & 0.0027 & 0.11 & 0.011 & 0.28 & 0.011 & 0.12 & 0.012 & 0.058 & 0.010 & 0.28 & 0.0090 \\
        
        & \ac{OOD}  & 2.98 & 0.041 & 5.42 & 0.16 & 7.87 & 0.12 & 20.12 & 0.25 & 28.57 & 2.40 & 5.47 & 4.16 & 5.47 & 1.21 & 13.91 & 5.53 \\
        
        & d$_2$ & 0.0091 & 1.7e-4 &  0.012 & 2.6e-4 & 0.049 & 0.0016 & 0.12 & 0.0017 & 0.12 & 0.0029 & 0.017 & 0.0021 & 0.040 & 0.0050 & 0.063 & 0.0058\\
        
        & \ac{MLP} & 0.059 & 9.8e-4 & 0.11 & 0.0027 & 0.32 & 0.0067 & 1.083 & 0.014 & 1.04 & 0.054 & 0.24 & 0.081 & 0.29 & 0.044 & 0.72 & 0.10\\
        \midrule
        \multirow{4}{*}{I$^2$-GNN} & d$_1$ &  0.0027 & 2.8e-4 & 0.027 & 8e-4 & 0.0054 & 5.8e-4 & 0.021 & 0.0023 & 0.065 & 0.0030 & 0.080 & 0.0079 & 0.035 & 0.0062 & 0.065 & 0.0032 \\
        
        & \ac{OOD}& 3.27 & 0.044 & 1.61 & 0.046 & 21.1 & 0.30 & 6.83 & 0.086 & 21.51 & 1.97 & 3.14 & 2.34 & 2.
        16 & 0.45 & 7.29 & 1.96\\
        
        & d$_2$ & 0.032 & 6.2e-4 & 0.028 & 5.9e-4 & 0.048 & 0.0015 & 0.30 & 0.0042 & 0.18 & 0.0057 & 0.070 & 0.0093 & 0.028 & 0.0031 & 0.12 & 0.011\\
        
        & \ac{MLP} & 0.2 & 0.0031 &  0.19 & 0.0048 & 0.91 & 0.015 & 1.59 & 0.021 & 1.2 & 0.032 & 0.22 & 0.045 & 0.19 & 0.026 & 0.37 & 0.066\\
        \bottomrule
    \end{tabular}
    }
\end{table}